%% file: main.tex
\documentclass[sigconf,screen]{acmart}
\input{setup}
\usepackage{multirow}

\AtBeginDocument{%
  \providecommand\BibTeX{{%
    \normalfont B\kern-0.5em{\scshape i\kern-0.25em b}\kern-0.8em\TeX}}}

\copyrightyear{2023}
\acmYear{2023}
\setcopyright{acmlicensed}\acmConference[KDD '23]{Proceedings of the 29th ACM SIGKDD Conference on Knowledge Discovery and Data Mining}{August 6--10, 2023}{Long Beach, CA, USA}
\acmBooktitle{Proceedings of the 29th ACM SIGKDD Conference on Knowledge Discovery and Data Mining (KDD '23), August 6--10, 2023, Long Beach, CA, USA}
\acmPrice{15.00}
\acmDOI{10.1145/3580305.3599447}
\acmISBN{979-8-4007-0103-0/23/08}

\author{Haruka Kiyohara}
\affiliation{Hanjuku-Kaso Co., Ltd.}
\email{kiyohara@hanjuku-kaso.com}

\author{Masatoshi Uehara}
\affiliation{Cornell University}
\email{mu223@cornell.edu}

\author{Yusuke Narita}
\affiliation{Yale University}
\email{yusuke.narita@yale.edu}

\author{Nobuyuki Shimizu}
\affiliation{Yahoo Japan Corporation}
\email{nobushim@yahoo-corp.jp}

\author{Yasuo Yamamoto}
\affiliation{Yahoo Japan Corporation}
\email{yasyamam@yahoo-corp.jp}

\author{Yuta Saito}
\affiliation{Cornell University}
\email{ys552@cornell.edu}

\begin{document}
\title[Off-Policy Evaluation of Ranking Policies under Diverse User Behavior]{Off-Policy Evaluation of Ranking Policies \\under Diverse User Behavior}


\renewcommand{\shortauthors}{Haruka Kiyohara et al.}

\begin{abstract}
\textit{Ranking} interfaces are everywhere in online platforms. There is thus an ever growing interest in their \textit{Off-Policy Evaluation} (OPE), aiming towards an accurate performance evaluation of ranking policies using logged data. A de-facto approach for OPE is \textit{Inverse Propensity Scoring} (IPS), which provides an unbiased and consistent value estimate. However, it becomes extremely inaccurate in the ranking setup due to its high variance under large action spaces. To deal with this problem, previous studies assume either independent or cascade user behavior, resulting in some ranking versions of IPS. While these estimators are somewhat effective in reducing the variance, all existing estimators apply a single universal assumption to every user, causing excessive bias and variance. Therefore, this work explores a far more general formulation where user behavior is diverse and can vary depending on the user context. We show that the resulting estimator, which we call \textit{Adaptive IPS} (AIPS), can be unbiased under any complex user behavior. Moreover, AIPS achieves the minimum variance among all unbiased estimators based on IPS. We further develop a procedure to identify the appropriate user behavior model to minimize the mean squared error (MSE) of AIPS in a data-driven fashion. Extensive experiments demonstrate that the empirical accuracy improvement can be significant, enabling effective OPE of ranking systems even under diverse user behavior.
\end{abstract}



\maketitle
\input{draft}

\bibliographystyle{ACM-Reference-Format}
\balance
\bibliography{ref.bib}

\input{appendix}

\end{document}

%% file: setup.tex
\usepackage{algorithm}
\usepackage{algorithmic}
\usepackage{bbding}

\usepackage{multicol,multirow}
\usepackage{bbding}
\usepackage{comment}
\usepackage{bm}
\usepackage{balance}

\DeclareMathOperator*{\argmin}{arg\,min}

\newcommand{\mE}{\mathbb{E}}
\newcommand{\mV}{\mathbb{V}}
\newcommand{\calD}{\mathcal{D}}
\newcommand{\calX}{\mathcal{X}}
\newcommand{\calA}{\mathcal{A}}

\newcommand{\mx}{\bm{x}}
\newcommand{\ma}{\bm{a}}
\newcommand{\mr}{\bm{r}}
\newcommand{\mc}{\bm{c}}
\newcommand{\tmc}{\tilde{\bm{c}}}

\newcommand{\ips}{\hat{V}_k^{\mathrm{IPS}} (\pi; \calD)}
\newcommand{\iips}{\hat{V}_k^{\mathrm{IIPS}} (\pi; \calD)}
\newcommand{\rips}{\hat{V}_k^{\mathrm{RIPS}} (\pi; \calD)}
\newcommand{\gips}{\hat{V}_k^{\mathrm{AIPS}} (\pi; \calD)}

\newcommand{\IPS}{\hat{V}_k^{\mathrm{IPS}}}
\newcommand{\AIPS}{\hat{V}_k^{\mathrm{AIPS}}}

\definecolor{dkred}{rgb}{0.8,0,0}
\definecolor{tickgreen}{rgb}{0,0.6,0}

\newcommand{\red}[1]{\textcolor{dkred}{#1}}

%% file: draft.tex
\section{Introduction}
\textit{Ranking} interfaces serve as a crucial component in many real-world search and recommender systems, where rankings (as actions) are often optimized through contextual bandit processes~\citep{mcinerney2020counterfactual, kiyohara2022doubly, dimakopoulou2019marginal}. As these ranking policies interact with the environment, they collect logged data valuable for \textit{Off-Policy Evaluation} (OPE)~\citep{saito2020open, saito2021counterfactual}. The goal of OPE is to accurately evaluate the performance of new policies using only the logged data without deploying them in the field, providing a safe alternative to online A/B testing~\citep{gilotte2018offline, kiyohara2021accelerating}.

A popular approach for OPE is \textit{Inverse Propensity Scoring} (IPS)~\citep{precup2000eligibility, strehl2010learning}, which provides an unbiased estimate of policy performance. Although several estimators are developed on top of IPS in standard OPE~\citep{wang2016optimal,su2020doubly}, they can severely degrade in large action spaces~\citep{saito2022off}. In particular, IPS becomes vulnerable when applied to ranking policies where the number of actions grows exponentially~\citep{li2018offline, mcinerney2020counterfactual}. An existing popular approach to deal with this variance issue is to make some assumptions about user behavior and define the corresponding IPS estimators~\citep{li2018offline,mcinerney2020counterfactual,kiyohara2022doubly}. For instance, \textit{Independent IPS}~ (IIPS)~\citep{li2018offline} assumes that a user interacts with the items in a ranking completely independently. In contrast, \textit{Reward interaction IPS} (RIPS)~\citep{mcinerney2020counterfactual} assumes that a user interacts with the items sequentially from top to bottom, namely the cascade assumption~\citep{guo2009efficient}. These estimators provide some variance reduction, while remaining unbiased when the corresponding assumption is satisfied.

Although this approach has been shown to improve IPS in some ranking applications~\citep{mcinerney2020counterfactual, kiyohara2022doubly}, a critical limitation is that all existing estimators model every user's behavior based on a single, universal assumption (such as independence and cascade). However, it is widely acknowledged that real user behavior can be much more diverse~\citep{xu2012incorporating, zhang2021constructing, mao2018constructing} and heterogeneous~\citep{zhang2022global}. With such diverse user behavior, the existing approach can suffer from significant bias and variance. For instance, consider a scenario with two groups of users, one following independent user behavior and the other following a cascade model. In this situation, IIPS is no longer unbiased, and RIPS is sub-optimal in terms of variance. An ideal strategy would arguably be to apply IIPS and RIPS to each group adaptively.

We thus explore a much more general formulation assuming that user behavior is sampled from a distribution conditional on the user context to capture possibly diverse behavior. On top of our general formulation, we propose a new estimator called \textit{Adaptive IPS} (AIPS), which applies different importance weighting schemes to different users based on their behavior, namely \textit{adaptive importance weighting}. We show that AIPS is unbiased for virtually any distribution of user behavior and that AIPS achieves the minimum variance among all unbiased IPS estimators. We also analyze the bias-variance tradeoff of AIPS in the case of unknown user behavior, which interestingly implies that the true behavior model may not result in an optimal OPE. Thus, we develop a strategy to \textit{optimize} the behavior model from the logged data in a way that minimizes the MSE of AIPS rather than merely trying to estimate the true behavior model. Experiments on synthetic and real-world data demonstrate that AIPS provides a significant gain in MSE over existing methods particularly when the user behavior is diverse.

Our contributions can be summarized as follows.
\begin{itemize}
    \item We propose a novel formulation and estimator for OPE of ranking policies capturing diverse user behavior.
    \item We show that AIPS is unbiased for any distributions of user behavior and that it achieves the minimum variance.
    \item We develop a non-parametric procedure to minimize the MSE of AIPS through optimizing (rather than estimating) the behavior model from the logged data.
    \item We empirically demonstrate that AIPS enables much more accurate OPE particularly under diverse user behavior.
\end{itemize}

\section{Preliminaries}
This section formulates OPE of ranking policies.

\subsection{Off-Policy Evaluation of Ranking Policies}
We use $\mx \in \calX \subseteq \mathbb{R}^d$ to denote a context vector (e.g., user demographics) and $\calA$ to denote a finite set of discrete actions. Let then $\ma= (a_1, a_2, \ldots, a_k, \ldots, a_K)$ denote a ranking action vector of length $K$ (e.g., a ranked list of songs). We call a function $\pi: \calX \rightarrow \Delta (\calA^K)$ a \textit{factored} policy. Given context $\mx$, it chooses an action at each position ($a_k$) independently, where $\pi(\ma \,|\, \mx) = \prod_{k=1}^K \pi(a_k \,|\, \mx)$ is the probability of choosing a specific ranking action $\ma$. In contrast, we call $\pi: \calX \rightarrow \Delta (\Pi_K(\calA))$ a \textit{non-factored} policy, where $\Pi_K(\calA)$ is a set of $K$-permutation of $\mathcal{A}$. Note that a \textit{factored} policy may choose the same action more than once in a ranking, whereas a \textit{non-factored} policy selects a ranking action without replacement (i.e., $\forall 1 \leq k < l \leq K, a_k \neq a_l$). In addition, let $\mr = (r_1, r_2, \ldots, r_k, \ldots, r_K)$ denote a reward vector with $r_k$ being a random reward observed at the $k$-th position (e.g., clicks, conversions, dwell time).

In OPE of ranking policies, we are interested in estimating the following \textit{policy value} of evaluation policy $\pi$ as a measure of its effectiveness~\citep{mcinerney2020counterfactual, kiyohara2022doubly}:
\begin{align}
    V(\pi) 
    :&= \mE_{p(\mx)\pi(\ma|\mx)}\left[ \sum_{k=1}^K \alpha_k q_k(\mx,\ma) \right] \notag \\
    &= \sum_{k=1}^K \alpha_k \underbrace{\mE_{p(\mx)\pi(\ma|\mx)}[q_k(\mx,\ma)]}_{V_k(\pi)}, \label{eq:value}
\end{align}
where $q_k(\mx,\ma) := \mE[r_k\,|\,\mx,\ma]$ is the position-wise expected reward function given context $\mx$ and ranking action $\ma$. $V_k(\pi)$ is the \textit{position-wise} policy value and $\alpha_k$ is a non-negative weight assigned to position $k$. Our definition of the policy value in Eq.~\eqref{eq:value} captures a wide variety of information retrieval metrics. For example, when $\alpha_k := 1/\log_2 (k+1)$, $V(\pi)$ becomes identical to the discounted cumulative gain (DCG)~\citep{jarvelin2002cumulated} under policy $\pi$. Throughout this paper, we focus on estimating the position-wise policy value $V_k(\cdot)$, as estimating $V(\cdot)$ is straightforward given an estimate of $V_k(\cdot)$.

For performing an OPE, we can leverage logged bandit data collected under the \textit{logging policy} $\pi_0$, i.e., $\calD := \{(\mx_i,\ma_i,\mr_i)\}_{i=1}^n$ where $\ma_i$ is a vector of discrete variables that indicate which ranking action is chosen by $\pi_0$ for individual $i$. $\mx_i$ and $\mr_i$ denote the context and reward vectors observed for $i$. To sum, a logged bandit dataset is generated in the following process:
\begin{align*}
  & \{(\mx_i,\ma_i,\mr_i)\}_{i=1}^n \sim \prod_{i=1}^n p(\mx_i) \pi_0 (\ma_i \,|\, \mx_i) p(\mr_i \,|\, \mx_i, \ma_i).
\end{align*}
Note that we assume that the logging policy provides full support over the ranking action space. The accuracy of an estimator $\hat{V}$ is measured by its MSE, i.e., $\mathrm{MSE} (\hat{V}) := \mE_{\calD} [ (V(\pi) - \hat{V}(\pi;\calD))^2 ]$, which can be decomposed into squared bias and variance of $\hat{V}$.

\subsection{Existing Estimators}
Here, we summarize some notable existing estimators for OPE in the ranking setup and their statistical properties.

\paragraph{Inverse Propensity Scoring} 
IPS uses the \textit{ranking-wise} importance weight to provide an unbiased and consistent estimate as follows.
\begin{align*}
    \ips := \frac{1}{n} \sum_{i=1}^n \frac{\pi(\ma_i\,|\,\mx_i)}{\pi_0(\ma_i\,|\,\mx_i)} r_{i,k}.
\end{align*}
IPS does not impose any particular user behavior model, and thus it is generally unbiased and consistent under standard identification assumptions. However, it suffers from extremely high variance when the action space ($|\calA^K|$ or $|\Pi_K(\calA)|$) is large~\citep{saito2022off}, which is particularly problematic in the ranking setup~\citep{swaminathan2017off, li2018offline, mcinerney2020counterfactual,kiyohara2022doubly}.

\paragraph{Independent IPS} 
IIPS assumes that a user interacts with the actions in a ranking independently, which is known as the independence assumption or item-position model~\citep{li2018offline}. This assumption posits that the reward observed at each position depends solely on the action chosen at that particular position, not on the other actions presented in the same ranking. Under this independence assumption, it is sufficient to condition only on $a_k$ to characterize the corresponding position-wise expected reward, i.e., $ q_{k} (\mx, \ma) = \mE [r_k \,|\,\mx, \red{a_k} ] $. Based on this assumption, IIPS defines the \textit{position-wise} importance weight as follows.
\begin{align*}
    \iips 
    &:= \frac{1}{n} \sum_{i=1}^n \frac{\pi(a_{i,k} \,|\, \mx_i)}{\pi_0(a_{i,k} \,|\, \mx_i)} r_{i,k}.
\end{align*}
where $\pi(a_k \,|\,\mx) := \sum_{\ma'} \pi(\ma' \,|\, \mx) \, \mathbb{I}\{a_k^{\prime} = a_k\}$ is the marginal action choice probability at position $k$ under policy $\pi$. IIPS substantially reduces the variance of IPS while remaining unbiased under the independence assumption. However, since the independence assumption is overly restrictive to describe real user behavior, IIPS often suffers from severe bias~\citep{mcinerney2020counterfactual, kiyohara2022doubly}.

\paragraph{Reward interaction IPS} 
RIPS leverages a weaker assumption called the cascade assumption, which assumes that a user interacts with the actions in a ranking sequentially from top to bottom~\citep{guo2009efficient}. Hence, the reward observed at each position ($r_k$) is influenced only by the actions observed at higher positions ($\ma_{1:k}$). Since the cascade model assumes that $r_k$ is independent of lower positions, it is sufficient to condition on $\ma_{1:k}$ to identify the position-wise expected reward, i.e., $q_{k} (\mx, \ma)  = \mE [r_{k} \,|\,\mx, \red{\ma_{1:k}} ]$. Based on this assumption, RIPS applies the \textit{top-k} importance weight as
\begin{align*}
    \rips 
    &:= \frac{1}{n} \sum_{i=1}^n \frac{ \pi(\ma_{i,1:k} \,|\, \mx_i)}{\pi_0(\ma_{i,1:k} \,|\, \mx_i)} r_{i,k}.
\end{align*}
where $\ma_{i,k_1:k_2}:=(a_{i,k_1}, a_{i,k_1+1}, \ldots, a_{i,k_2})$. RIPS is unbiased under the cascade assumption, while reducing the variance of IPS~\citep{mcinerney2020counterfactual}. However, when the cascade assumption does not hold true, it may produce a large bias. Furthermore, RIPS can suffer from high variance when the ranking size is large~\citep{kiyohara2022doubly}.

\paragraph{\textbf{Limitation of the existing estimators.}}
We have so far seen that existing estimators have tried to control the bias-variance tradeoff by leveraging some assumptions on user behavior -- a stronger assumption reduces the variance more but introduces a larger bias. Although this approach has shown some success, a critical limitation is that existing estimators apply a single universal assumption to the entire population, while real user behavior can often be much more diverse and heterogeneous~\citep{zhang2022global,xu2012incorporating,zhang2021constructing,mao2018constructing}. In such realistic scenarios, imposing a single assumption can result in highly sub-optimal estimations. For example, a strong assumption (e.g., independence) produces a large bias in a subpopulation following more complex behavior models, while a weak assumption (e.g., cascade) produces unnecessary variance in another subpopulation following simpler behavior models. This limitation motivates the development of a new estimator that can better exploit the potentially diverse and heterogeneous user behavior to substantially improve OPE of ranking policies.

\begin{table}[t]
    \centering
    \caption{Correspondence among user behavior assumptions, estimators, and relevant set of actions.}
    \label{tab:gips}
    \begin{tabular}{c|c|cc}
        \toprule
        assumption & estimator & relevant actions $\Phi_k(\ma,\mc)$ \\ \midrule \midrule
        no assumption & IPS & $\ma$ \\
        cascade & RIPS & $\ma_{1:k}$ \\
        independence & IIPS &  $a_k$ \\ 
        \midrule
        adaptive &  AIPS (ours)  & $\Phi_k(\ma,\mc), \; \mc \sim p(\cdot|\mx)$ \\
        \bottomrule
    \end{tabular}
\end{table}

\section{The Adaptive IPS Estimator}
Our key idea in deriving a new estimator is to take into account various user behaviors by refining the typical formulation of OPE of ranking systems. More specifically, here we introduce an \textit{action-reward interaction} matrix denoted by $\mc \in \{0,1\}^{K \times K}$ whose $(k,l)$ element ($c_{k,l}$) indicates whether $r_k$ is affected by $a_{l}$. Given $\mc$, the position-wise expected reward can be expressed as follows.
\begin{align*}
    q_{k} (\mx, \ma, \mc)  = \mE [r_{k} \,|\, \mx, \red{\Phi_k(\ma, \mc)} ],
\end{align*}
where $\Phi_k(\ma, \mc) := \{ a_{l} \in \calA \,|\, c_{k,l}=1 \}$ is a set of \textit{relevant} or \textit{sufficient} actions needed to identify the expected reward function at the $k$-th position. The elements of the matrix are considered to be sampled from some unknown probability distribution $p(\mc \,|\, \mx)$, which is conditioned on the context $\mx$ to capture potentially diverse and heterogeneous user behavior. Table~\ref{tab:gips} describes how our formulation generalizes the assumptions used by existing estimators.

Leveraging the action-reward interaction matrix, our \textit{Adaptive IPS} (AIPS) estimator is defined as
\begin{align*}
    \gips := \frac{1}{n} \sum_{i=1}^n \displaystyle \frac{\pi(\Phi_k(\ma_i, \mc_i)\,|\,\mx_i)}{\pi_0(\Phi_k(\ma_i, \mc_i)\,|\,\mx_i)} r_{i,k}.
\end{align*}
At a high level, AIPS applies \textit{adaptive} importance weighting based on the context-aware behavior model $\mc$. Specifically, when estimating the position-wise policy value at the $k$-th position, AIPS considers only the actions that affect the reward observed at that particular position ($\Phi_k(\ma_i, \mc_i)$) to define the importance weight. In this way, AIPS is able not only to deal with potential bias due to diverse user behavior but also to avoid producing unnecessary variance. The following sections show how AIPS enables a much more effective OPE of ranking policies compared to existing estimators.

\begin{table*}[t]
    \centering
    \caption{A toy example illustrating the possible benefit of strategic variance reduction with an incorrect behavior model. AIPS with an incorrect (but optimized) behavior model produces much smaller variance while introducing some small bias, resulting in a smaller MSE than AIPS with the true behavior model.}
    \label{tab:strategic-variance-reduction}
    \vspace{-2mm}
    \scalebox{1.1}{
    \begin{tabular}{c|ccc}
        \toprule
         & bias & variance & \textbf{MSE} (= bias$^2$ + variance) \\ \midrule
        AIPS with the true behavior model $\mc$ & 0.0 & 0.5 & \textbf{0.50} ($\,= (0.0)^2 + 0.5$ )\\
        AIPS with an incorrect (but optimized) behavior model $\hat{\mc}$ & 0.1 & 0.3 &  \textbf{0.31} ($\,=(0.1)^2 + 0.3$)\\
        \bottomrule
    \end{tabular}}
\end{table*}

\subsection{Theoretical Analysis}
This section provides some key statistical properties of AIPS assuming that the user behavior model $\mc$ is observable. Then, we analyze the bias of AIPS when using an estimated user behavior model $\hat{\mc}$. Finally, we present an algorithm to \textit{optimize} the behavior model in a way that minimizes the MSE of the resulting estimator. Note that all proofs omitted from the main text are provided in Appendix~\ref{app:proof}.

First, we show that AIPS can be unbiased under any (context-dependent) distribution of user behavior $p(\mc \,|\, \mx)$.
\begin{proposition} \label{prop:unbiased}
If the user behavior model $\mc$ is observed, AIPS is unbiased, i.e., $\mE_{\calD}[\gips] = V_k(\pi)$ for any $\pi$ and $p(\mc \,|\, \mx)$. 
\end{proposition} 

Proposition~\ref{prop:unbiased} suggests that AIPS is unbiased in a far more general situation about user behavior compared to that of existing work, which relies on a particular behavior model. Next, we show that the variance reduction of AIPS from IPS can be substantial.

\begin{theorem} \label{thrm:variance}
(Variance Reduction of AIPS over IPS)
Compared to IPS, AIPS reduces the variance by the following amount.
\begin{align*}
     & n \left(\mV_{\calD}(\IPS (\pi; \calD) ) - \mV_{\calD}(\AIPS (\pi; \calD)) \right)  \\
     & \!\! = \mE \Bigg[ \left( \frac{\pi(\Phi_k(\ma,\mc)\,|\,\mx)}{\pi_0(\Phi_k(\ma,\mc)\,|\,\mx)} \right)^2
      \mV_{\Phi_k^c(\ma,\mc)} \left(  \frac{\pi(\Phi_k^c(\ma,\mc)\,|\,\mx,\Phi_k(\ma,\mc))}{\pi_0(\Phi_k^c(\ma,\mc)\,|\,\mx,\Phi_k(\ma,\mc))} \right) \\
     & \qquad \qquad \qquad \qquad \qquad \qquad \qquad \qquad \quad \cdot \mE \left[ r_k^2 \,|\, \mx,\Phi_k(\ma,\mc) \right] \Bigg],
\end{align*}
where the outer expectation is taken over $p(\mx)p(\mc|\mx)\pi_0(\Phi_k(\ma,\mc)|\mx)$, and $\Phi_k^c(\ma,\mc)$ is the complement of $\Phi_k(\ma,\mc)$.
\end{theorem}

Theorem~\ref{thrm:variance} ensures that AIPS always provides a non-negative variance reduction over IPS. Moreover, Theorem~\ref{thrm:variance} suggests that variance reduction becomes substantial when $\mc$ is sparse and the importance weight about irrelevant actions ($\Phi_k^c(\ma,\mc)$) is large.

The following also shows that AIPS achieves the minimum variance among all IPS-based unbiased estimators.
\begin{theorem} \label{thrm:minimum}
(Variance Optimality of AIPS)
Let $$\hat{V}_k(\pi; \calD, \tilde{\mc}) :=  \frac{1}{n} \sum_{i=1}^n \frac{\pi(\Phi_k(\ma_i,\tilde{\mc})\,|\,\mx_i)}{\pi_0(\Phi_k(\ma_i,\tilde{\mc})\,|\,\mx_i)} r_{i,k},$$ 
so that $ \mathbb{E} [\hat{V}_k(\pi; \calD, \tilde{\mc})] = V_k(\pi) $.
Then, for any $\tilde{\mc}$ (s.t. $\mc \subseteq \tilde{\mc}$) and $\pi$, we have
\begin{align*}
    \mV_{\calD}\big(\hat{V}_k^{\mathrm{AIPS}}(\pi; \calD) \big) \le \mV_{\calD}\big(\hat{V}_k(\pi; \calD, \tilde{\mc})\big).
\end{align*}
\end{theorem}
Theorem~\ref{thrm:minimum} ensures that AIPS guarantees the minimum variance among all unbiased IPS estimators under any distribution of user behavior, suggesting that AIPS is the optimal unbiased estimator.\footnote{Note that AIPS may not be optimal if we also take some \textit{biased} estimators into consideration. This motivates our idea of intentionally leveraging an incorrect behavior model to further improve the MSE, which is the sum of the squared bias and variance.}

Although we have shown above that AIPS can exhibit favorable statistical properties under general user behavior, it should be noted that we currently assume that the true user behavior $\mc$ is observable. Since this is generally not the case, the following investigates the bias of AIPS when given is an estimated user behavior $\hat{\mc}$.
\begin{theorem} \label{thrm:bias}
(Bias of AIPS with an estimated user behavior)
When an estimated user behavior $\hat{\mc}$ is used, AIPS has the following bias.
\begin{align*}
    \mathrm{Bias}(\hat{V}_k^{\mathrm{AIPS}}; \hat{\mc}) 
    &= \mE_{p(\mx)p(\mc|\mx)\pi(\ma|\mx)} \left[ \left( \Delta w_k(\ma, \mc, \hat{\mc}) - 1 \right)  q_k(\mx,\ma,\mc) \right],
\end{align*}
where
\begin{align*}
    \Delta w_k(\ma, \mc, \hat{\mc}) := \frac{\pi_0(\Phi_k(\ma,\mc) \setminus \Phi_k(\ma,\hat{\mc}) \,|\, \mx,\Phi_k(\ma,\hat{\mc}))}{\pi(\Phi_k(\ma,\mc) \setminus \Phi_k(\ma,\hat{\mc}) \,|\, \mx,\Phi_k(\ma,\hat{\mc}))}.
\end{align*}
\end{theorem}
Theorem~\ref{thrm:bias} suggests that AIPS remains unbiased when the true model is a subset of the estimated model ($\mc \subseteq \hat{\mc}$). Furthermore, we can see that the bias of AIPS is characterized by the overlap between the true $\mc$ and estimated user behavior $\hat{\mc}$, i.e., when there is a large overlap between $\mc$ and $\hat{\mc}$, the cardinality of $\Phi_k(\ma,\mc) \setminus \Phi_k(\ma,\hat{\mc})$ becomes small, resulting in a smaller bias for AIPS.

\paragraph{\textbf{Controlling the bias-variance tradeoff}} 
Theorems~\ref{thrm:variance} and~\ref{thrm:bias} suggest that the bias-variance tradeoff of AIPS is mainly characterized by $\hat{\mc}$.\footnote{Note that the variance of AIPS with an estimated behavior model can immediately be obtained by replacing $\mc$ with $\hat{\mc}$ in Theorem~\ref{thrm:variance}.} When $\hat{\mc}$ is dense, the bias of the resulting AIPS estimator will be very small, but the variance can be high. On the other hand, a sparse $\hat{\mc}$ can substantially reduce the variance of AIPS while introducing some bias. This suggests that true user behavior $\mc$ may not necessarily minimize the MSE of AIPS, and that there exists an interesting strategy to intentionally utilize an incorrect model $\hat{\mc}$ to further improve the accuracy of the downstream estimation. Table~\ref{tab:strategic-variance-reduction} provides a toy example illustrating a situation where AIPS with an incorrect behavior model can achieve a lower MSE than that with the true behavior model. In this example, the minimum variance among all unbiased estimators is 0.5, which is achieved by the true behavior model as per Theorem~\ref{thrm:minimum}. However, a lower MSE can be realized by intentionally using an incorrect (overly sparse) model. This is because we can gain a large variance reduction (-0.2) by allowing only a small squared bias (+0.01), and hence using the true user behavior model does not result in the optimal MSE of AIPS. Therefore, instead of discussing how to \textit{estimate} the true user behavior, the following section describes a data-driven approach to \textit{optimize} $\hat{\mc}$ in a way that minimizes the MSE of AIPS.

\begin{figure}[tb]
    \centering
    \includegraphics[clip, width=0.925\linewidth]{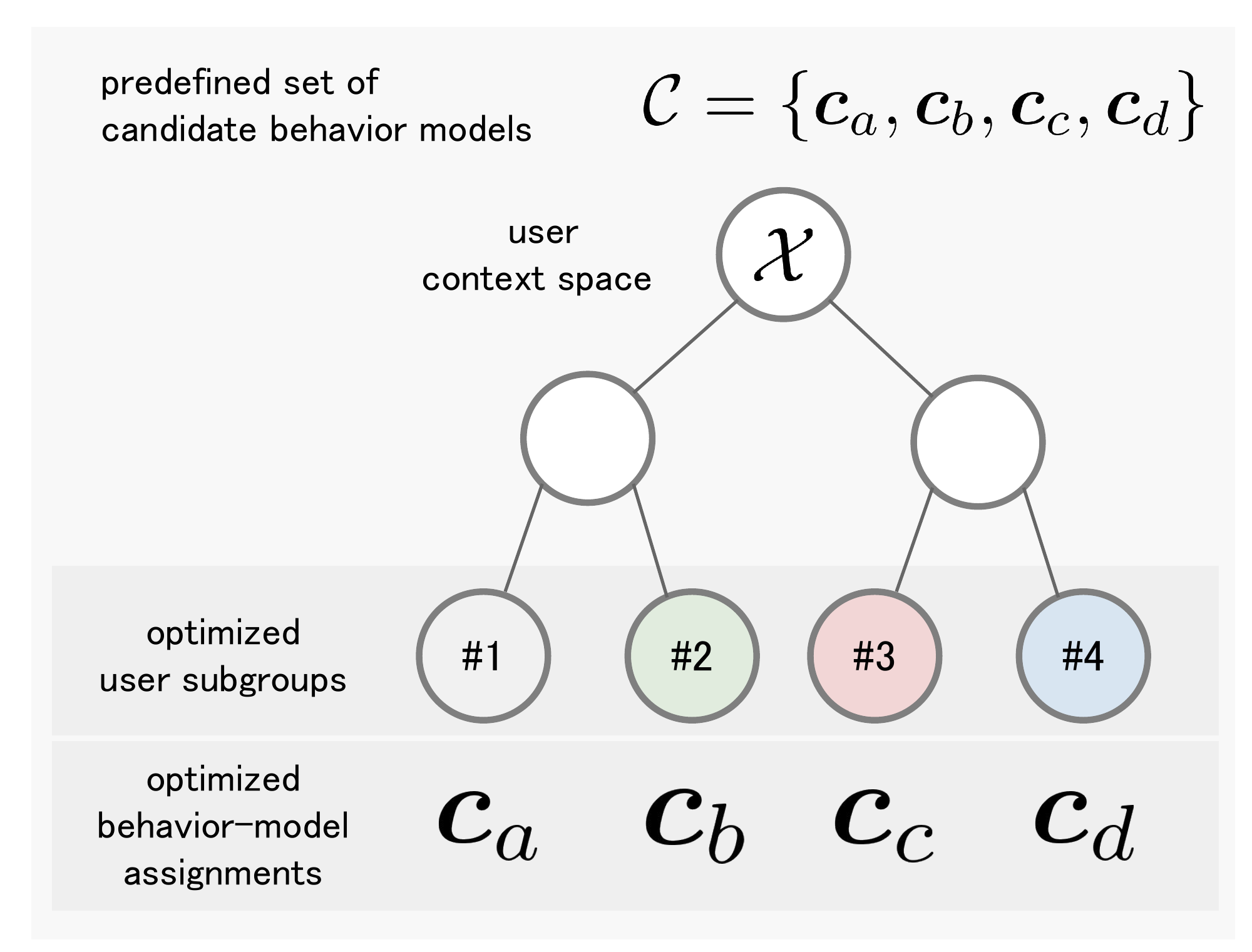}
    \caption{Tree-based optimization of user behavior model, which optimizes the partition in the context space and assignments of the user behavior model of each subgroup (from \#1 to \# 4) so the resulting MSE of AIPS is minimized.}
    \label{fig:ubsi}
\end{figure}

\subsection{\textit{Optimizing} the User Behavior Model} \label{sec:ubso}
Our goal here is to \textit{optimize} the context-aware user behavior model $\hat{\mc}(\mx)$ to minimize the MSE of AIPS, rather than merely identifying the true model. We achieve this by optimizing the behavior model at a \textit{subgroup level}, inspired by the subgroup identification techniques proposed in treatment effect estimation~\citep{athey2016recursive,keramati2022identification}.

Due to its flexibility in handling even non-differentiable objectives, we employ a non-parametric tree-based model to optimize the partition in the context space (user subgroups) and the assignments of user behavior models to each subgroup simultaneously. Specifically, we recursively partition the feature space ($\calX$) and assign an appropriate behavior model to each node in a way that minimizes the MSE of AIPS. This means that we follow the classification and regression tree (CART) algorithm~\citep{breiman2017classification} and set the MSE of AIPS as its objective function.
More specifically, we first define a candidate set of user behavior models $\mathcal{C} := \{ \hat{\mc}_0, \ldots, \hat{\mc}_{m} \}$.\footnote{In general, set of candidate models $\mathcal{C}$ should be as large as possible, but a larger $\mathcal{C}$ might be infeasible due to intensive computation. In experiments, we show that AIPS remains effective even when $\mathcal{C}$ is not large if we include a diverse set of models so that the tree model can find an appropriate behavior model for each user subgroup.} Then, at each parent node $l \subseteq \mathcal{X}$, the tree partitions it into child nodes ($l_{(l^{\ast})},l_{(r^{\ast})} \subseteq l$ where $l_{(l^{\ast})} \bigcup l_{(r^{\ast})} = l$ and $l_{(l^{\ast})} \bigcap l_{(r^{\ast})} = \emptyset$) and assigns behavior models $(\hat{\mc}_{(l^{\ast})}, \hat{\mc}_{(r^{\ast})} \in \mathcal{C})$ to these nodes by the following criterion.
\begin{align}
    (l_{(l^{\ast})}, l_{(r^{\ast})}, \hat{\mc}_{(l^{\ast})}, \hat{\mc}_{(r^{\ast})}) 
    &:= \argmin_{(l_{(l)}, l_{(r)}, \hat{\mc}_{(l)}, \hat{\mc}_{(r)})} \, \widehat{MSE}(\hat{\mc}_{(l)}, \hat{\mc}_{(r)}; l_{(l)}, l_{(r)})
\end{align}
where $\hat{\mc}_{(l)} \in \mathcal{C}$ is a candidate behavior model assigned to node $l$, and $\widehat{MSE}(\hat{\mc}_{(l)}, \hat{\mc}_{(r)}; l_{(l)}, l_{(r)})$ is an estimated MSE when $\hat{\mc}_{(l)},\hat{\mc}_{(r)} \in \mathcal{C}$ are assigned to the left and right nodes, respectively. Algorithm~\ref{algo:ubsi} in the appendix provides the complete optimization procedure.

Compared to existing subgroup identification procedures~\citep{athey2016recursive, keramati2022identification}, our algorithm is unique in that it directly optimizes the MSE in OPE rather than minimizing some prediction loss for treatment effect estimation. It is important to note that our algorithm is agnostic to the method used for estimating the MSE. For example, we can estimate the MSE by following existing methods from~\citep{su2020doubly} or~\citep{udagawa2022policy}. We thus consider the MSE estimation task as an independent research topic and do not propose specific approaches to estimate the MSE from the logged data. Instead, our experiments will demonstrate that AIPS with our data-driven procedure for behavior model optimization performs reasonably well across a variety of experiment settings, even with a noisy MSE estimate and with an MSE estimated via an existing method from~\citet{udagawa2022policy} that uses only the observed logged data.

\section{Synthetic Experiments} \label{sec:synthetic_experiment}
This section empirically compares the proposed estimator with existing estimators (IPS, IIPS, and RIPS) on synthetic ranking data. Our experiment is implemented on top of \textit{OpenBanditPipeline}~\citep{saito2020open}\footnote{\href{https://github.com/st-tech/zr-obp}{https://github.com/st-tech/zr-obp}}, a modular Python package for OPE. Our experiment code is available at \href{https://github.com/aiueola/kdd2023-aips}{https://github.com/aiueola/kdd2023-aips}. Other experiment details and additional results are provided in Appendix~\ref{app:experiment}.

\subsection{Setup}

\paragraph{\textbf{Basic setting}} 
To generate synthetic data, we randomly sample five-dimensional context ($d=5$) from the standard normal distribution. Then, for each position $k$, we sample continuous rewards from a normal distribution as $r_k \sim \mathcal{N}(q_k(\mx,\ma,\mc), \sigma^2)$, where we use $\sigma = 0.5$. The following describes how to define the expected reward function $q_k(\mx,\ma,\mc)$ and user behavior distribution $p(\mc \,|\, \mx)$.

\paragraph{\textbf{Position-wise expected reward function}}
Following~\citet{kiyohara2022doubly}, we first define the following position-wise \textbf{base} reward function $\tilde{q}_k(\mx, a_k)$, which depends only on the action presented at the corresponding position ($a_k$) rather than the entire ranking.
\begin{align*}
    \tilde{q}_k(\mx, a_k) =\theta_{a_k}^{\top} \mx + b_{a_k},
\end{align*}
where $\theta_{a_k}$ is a parameter vector sampled from the standard normal distribution, and $b_{a_k}$ is a bias term that corresponds to action $a_k$.

Then, we define the position-wise expected reward function given a particular user behavior model $\mc$ as follows.
\begin{align*}
    q_k(\mx, \ma, \mc) = c_{k,k} \, \tilde{q}_k(\mx, a_k) + \sum_{l \neq k} c_{k,l} \, \mathbb{W}(a_k, a_{l})
\end{align*}
where $c_{k,l} \in \{0, 1\}$ is the $(k,l)$ element of $\mc$, which indicates whether $a_{l}$ affects $r_k$. $\mathbb{W}$ is a $|\calA| \times |\calA|$ matrix whose elements are sampled from a uniform distribution with range $[0,1]$. This matrix defines how the co-occurrence of a pair of actions affects $q_k(\mx,\ma,\mc)$. 

\paragraph{\textbf{Distribution of user behavior}}
Next, the following defines the three basic user behavior models used in existing work~\citep{mcinerney2020counterfactual, kiyohara2022doubly}.
\begin{itemize}
    \item \textbf{standard (S)}: $\mc_{S} (k,l) = 1, \forall l \in [K].$
    \item \textbf{cascade (C)}: $\mc_{C} (k,l) = 1, \forall l \leq k, \text{and}
    \, \mc_{C} (k,l) = 0, \mathrm{otherwise}.$
    \item \textbf{independence (I)}: $\mc_{I} (k,k) = 1 \text{and}
    \, \mc_{I} (k,l) = 0, \mathrm{otherwise}.$
\end{itemize}
for each $k \in [K]$. To introduce more diverse behaviors beyond the above basic models, we define the following \textit{h-neighbor perturbation}:
\begin{align*}
    & c_{\textit{neighbor},h}(k,l) = 1, \forall |l - k| \leq h, \\
    & \qquad \text{and} \; c_{\textit{neighbor},h}(k,l) = 0, \text{otherwise}.
\end{align*}
where $h$ is the number of neighboring items that perturb the basic model.
By applying this perturbation to the basic models, we define the following more complex behavior models.
\begin{itemize}
    \item \textbf{C1}: $\mc_{C1} (k,l) = \mc_{C}(k,l) + c_{\textit{neighbor},1}(k,l)$
    \item \textbf{C2}: $\mc_{C2} (k,l) = \mc_{C}(k,l) + c_{\textit{neighbor},2}(k,l)$
    \item \textbf{I1}: $\mc_{I1} (k,l) = \mc_{I}(k,l) + c_{\textit{neighbor},1}(k,l)$
\end{itemize}

We also define two additional user behaviors by applying \textit{random perturbation} to the independence model as follows.
\begin{itemize}
    \item \textbf{R3}: $\mc_{R3} (k,l) = \mc_{I}(k,l) + c_{\textit{random},3}(k,l) \mathbb{I}\{k \neq l\}$
    \item \textbf{R6}: $\mc_{R6} (k,l) = \mc_{I}(k,l) + c_{\textit{random},6}(k,l) \mathbb{I}\{k \neq l\}$
\end{itemize}
where $c_{\textit{random},h}(k,\cdot)=1$ only for randomly chosen $h$ positions for each $k \in [K]$.

To study how the estimators work under diverse and heterogeneous user behaviors, we use \textbf{\{S, R6, R3, C2, C1, I1\}} and sample them from the following distribution given a user context:
\begin{align*}
    p(\mc_z\,|\,\mx) 
    := \operatorname{softmax} (\lambda_z \cdot |\theta_z^{\top} \mx|)
    = \frac{\exp(\lambda_z \cdot |\theta_z^{\top} \mx|)}{\sum_{z'} \exp(\lambda_{z'} \cdot |\theta_{z'}^{\top} \mx|)},
\end{align*}
where $z \in \{\mathrm{S, R6, R3, C2, C1, I1}\}$ is the index of each user behavior. $\theta_z$ is a parameter vector sampled from the standard uniform distribution, and $\lambda_z$ is some weight parameter. By assigning different values of $\lambda_z$ to different user behaviors, we can control the distribution of user behavior. In particular, we define $\lambda_z$ as follows.
\begin{align*}
    \lambda_z := \exp( (2 \delta - 1) \cdot \gamma_z ),
\end{align*}
where $\gamma_z$ is some coefficient value defined for each user behavior as $\{ \gamma_{\mathrm{S}}, \gamma_{\mathrm{R6}}, \gamma_{\mathrm{R3}}, \gamma_{\mathrm{C2}}, \gamma_{\mathrm{C1}}, \gamma_{\mathrm{I1}}
\} = \{ 1.5, 0.9, 0.3, -0.3, -0.9, -1.5\} $, and $\delta \in [0,1]$ is an experiment parameter called the ``user behavior distribution parameter'', which controls the entropy of the behavior distribution. For example, all user behavior will be uniformly distributed when $\delta=0.5$, as $\gamma_z = 0, \forall z$. In contrast, $\delta < 0.5$ samples user behaviors having negative values of $\gamma_z$ more frequently, while $\delta > 0.5$ prioritizes those having positive values of $\gamma_z$. In particular, under our definition of $\{\gamma_z\}$, a smaller value of $\delta$ leads to simpler user behavior, while a larger value leads to more complex behavior in general.\footnote{Figures~\ref{fig:variation_setting} and~\ref{fig:stochasticity_setting} in Appendix~\ref{app:experiment} show how changes in the value of distribution parameters ($\delta$ and $\lambda$) control the distribution of user behavior $p(\mc_z\,|\,\mx)$.}

\paragraph{\textbf{Logging and evaluation policies}}
We define a factored logging policy to generate synthetic logged data as follows.
\begin{align*}
    \pi_0(\ma \,|\, \mx) = \prod_{k=1}^K \pi_0(a_k \,|\, \mx) = \prod_{k=1}^K \frac{\exp(f_0(\mx, a_k))}{\sum_{a'\in\calA} \exp(f_0(\mx, a'))},
\end{align*}
where $f_0(\mx, a) = \theta_{a}^{\top} \mx + b_{a}$. We sample $\theta_{a}$ and $b_{a}$ from the standard uniform distribution. Then, we define the evaluation policy by applying the following transformation to the logging policy.
\begin{align}
    \pi(\ma \,|\, \mx) = \prod_{k=1}^K \left( (1 - \epsilon) \, \mathbb{I} \bigl\{ a_k = \argmin_{a^{\prime} \in \calA} f_0(\mx, a^{\prime}) \bigr\} + \epsilon / |\calA| \right),
\end{align}
where $\epsilon \in [0, 1]$ is an experiment parameter that determines the stochasticity of $\pi$. Specifically, a small value of $\epsilon$ leads to a near-deterministic policy, while a large value leads to a near-uniform policy. We use $\epsilon=0.3$ throughout our synthetic experiment.

\paragraph{\textbf{Compared estimators}}
We compare AIPS against IPS, RIPS, and IIPS. We also report the results of AIPS (true), which uses the true user behavior $\mc$ and thus is infeasible in practice. However, this provides a useful reference to investigate the effectiveness of our strategic variance reduction method from Section~\ref{sec:ubso}. 

Note that AIPS uses the following surrogate MSE as the objective function when performing user behavior optimization.
\begin{align*}
    \widehat{MSE} \big(\hat{V}^{\mathrm{AIPS}}_k(\pi; \calD, \tilde{\mc}) \big) 
    &= \widehat{Bias} \big(\hat{V}^{\mathrm{AIPS}}_k(\pi; \calD, \tilde{\mc}) \big)^2 + \hat{\mV} \big(\hat{V}^{\mathrm{AIPS}}_k(\pi; \calD, \tilde{\mc}) \big),
\end{align*}
where $\hat{\mV}(\cdot)$ is the sample variance. To control the accuracy of the bias estimation, we use its noisy estimate $\widehat{Bias}$. Specifically, we first estimate the bias based on an on-policy estimate of the policy value, which is denoted as $\widehat{Bias}_\mathrm{on} (\cdot)$, and then add some Gaussian noise as $\widehat{Bias} \sim \mathcal{N}(\widehat{Bias}_\mathrm{on}, \sigma_{\Delta}^2)$ where $\sigma_{\Delta} = 0.3 \times |\widehat{Bias}_{\mathrm{on}}|$. By doing so, we can simulate a practical situation where AIPS relies on some noisy estimate of MSE. This procedure also enables us to evaluate the robustness of AIPS to the varying accuracies of MSE estimation, as demonstrated in Appendix~\ref{app:experiment}.

\begin{figure*}[!htb]
\begin{minipage}[c]{0.99\hsize}
\centering
\scalebox{0.95}{
\begin{tabular}{c}
\begin{minipage}{0.60\hsize}
\begin{center}
\includegraphics[width=1.05\linewidth]{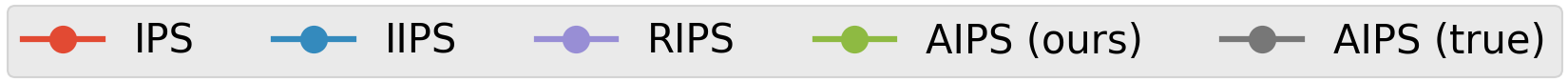}
\end{center}
\end{minipage}
\\
\\
\begin{minipage}{0.99\hsize}
    \begin{center}
        \includegraphics[clip, width=1.0\linewidth]{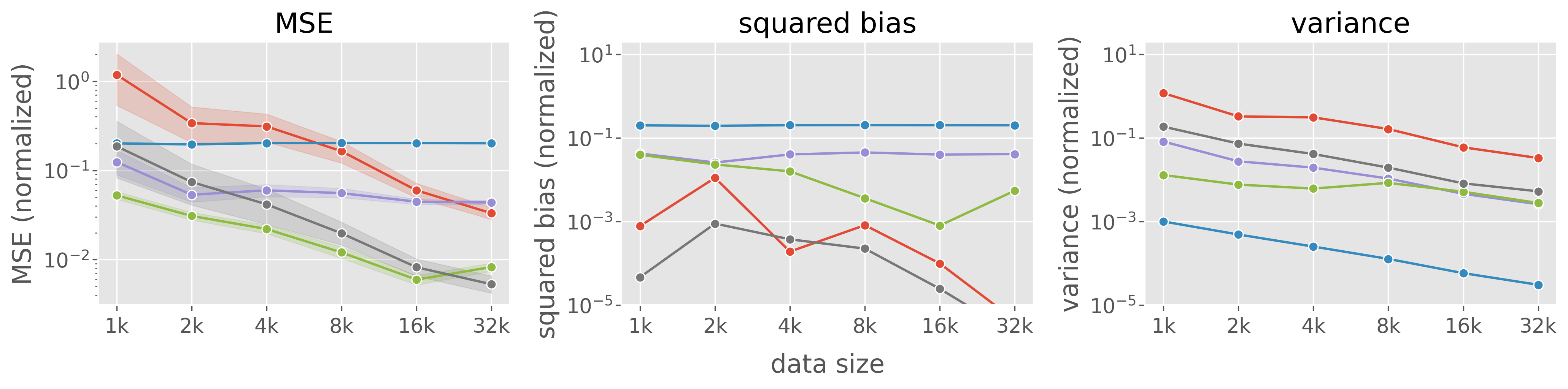}
        \vspace{-7mm}
        \caption{Comparison of the estimators' MSE (normalized by the true value $V(\pi)$) with varying data sizes ($n$)}
        \label{fig:data_size}
        \vspace{3mm}
    \end{center}
\end{minipage}
\\ 
\\
\begin{minipage}{0.99\hsize}
    \begin{center}
        \includegraphics[clip, width=1.0\linewidth]{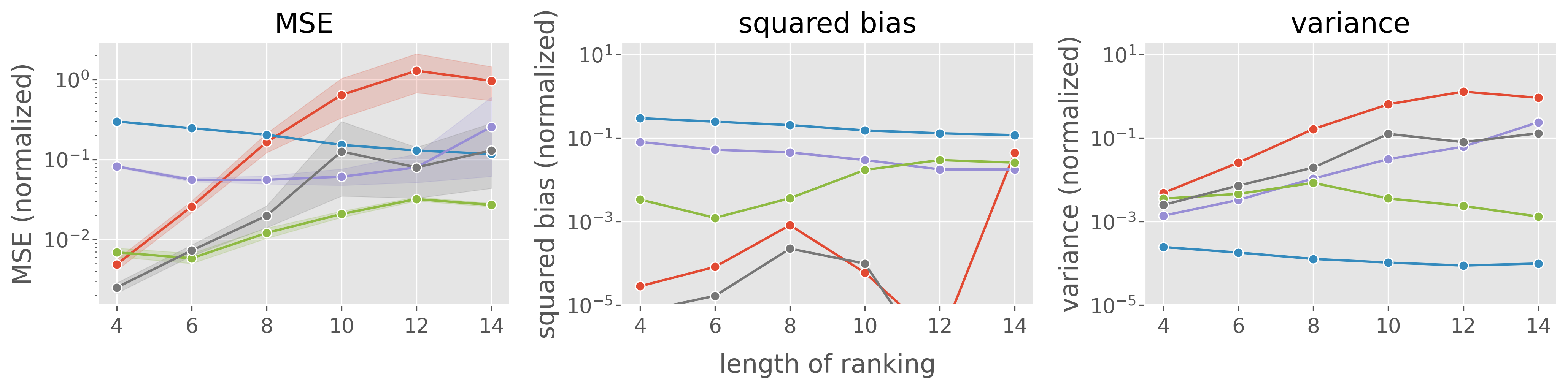}
        \vspace{-7mm}
        \caption{Comparison of the estimators' MSE (normalized by the true value $V(\pi)$) with varying lengths of ranking ($K$)}
        \label{fig:slate_size}
        \vspace{3mm}
    \end{center}
\end{minipage}
\\
\\
\begin{minipage}{0.99\hsize}
    \begin{center}
        \includegraphics[clip, width=1.0\linewidth]{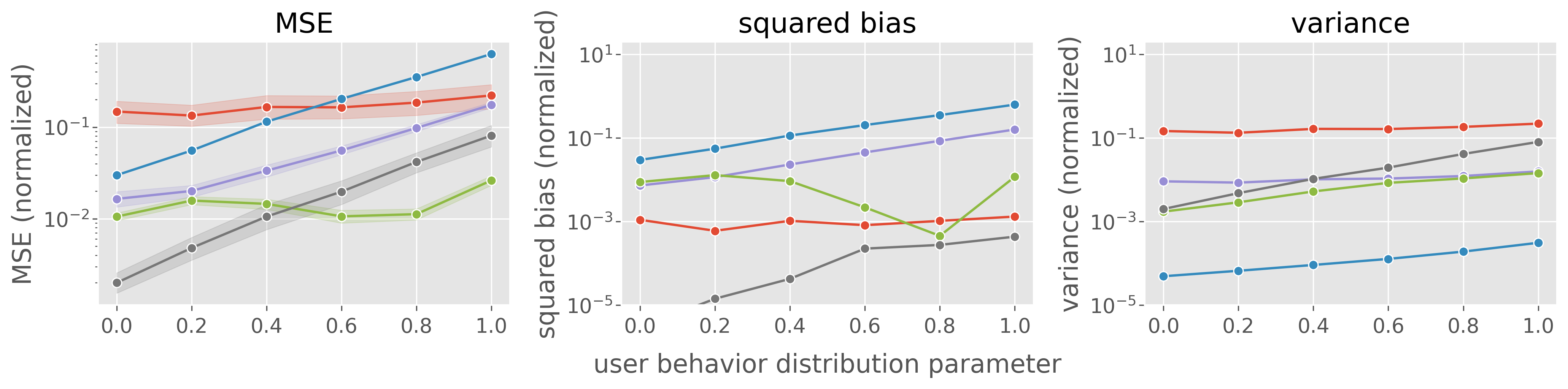}
        \vspace{-7mm}
        \caption{Comparison of the estimators' MSE (normalized by the true value $V(\pi)$) with varying behavior distributions ($\delta$)}
        \label{fig:user_behavior_variation}
    \end{center}
\end{minipage}
\\
\\
\end{tabular}
}
\end{minipage}
\end{figure*}

\begin{figure*}[tb]
    \centering
    \begin{minipage}[b]{0.40\linewidth}
    \centering
    \includegraphics[clip, width=0.95\linewidth]{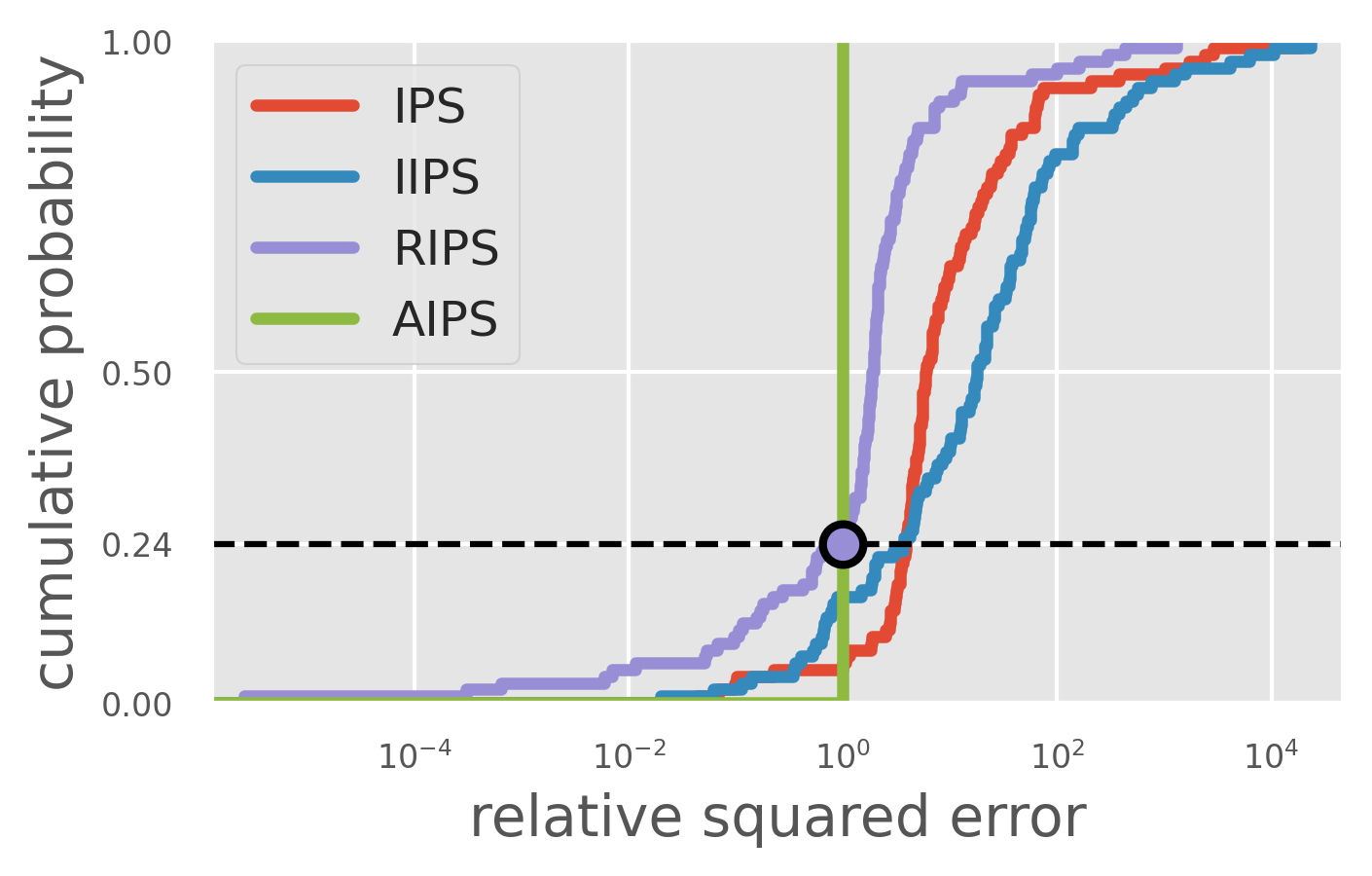}
    \label{fig:cdf}
    \end{minipage}
    \begin{minipage}[b]{0.40\linewidth}
    \centering
    \includegraphics[clip, width=0.95\linewidth]{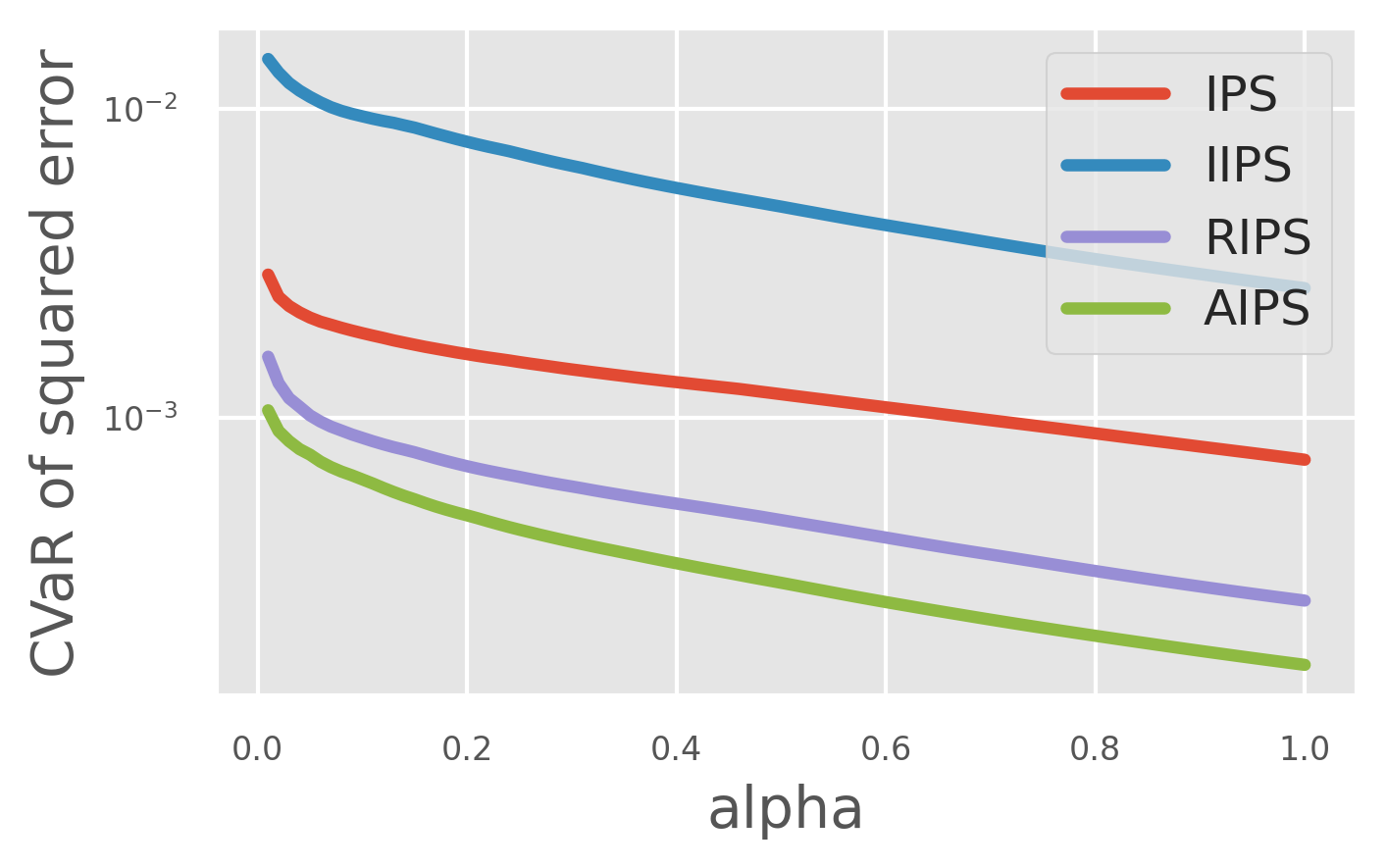}
    \label{fig:cvar}
    \end{minipage}
    \vspace{-2mm}
    \begin{minipage}[b]{0.95\linewidth}
    \caption{Estimators' performance comparison in the real-world experiment:
    (Left) Cumulative distribution function (CDF) of the estimators' squared error (relative to that of AIPS). 
    (Right) Conditional Value at Risk (CVaR) of the estimators' squared error with varying values of $\alpha$.}
    \label{fig:real}
    \end{minipage}
\end{figure*}

\subsection{Results and Discussion}
We run the OPE simulations 1000 times with different random seeds. We report the MSE, bias, and variance of the estimators normalized by the true policy value $V(\pi)$. Note that we use $n=8\mathrm{k}$, $K=8$, and $\delta=0.6$ as default experiment parameters.\footnote{We use $1\mathrm{k}, 2\mathrm{k}, \ldots$ to denote $1000, 2000, \ldots$.} In all figures, the solid lines indicate the performance metrics averaged over the simulation runs and the shaded regions show their 95\% confidence intervals.

\paragraph{\textbf{RQ (1): How do the estimators perform with varying data sizes?}}
Figure~\ref{fig:data_size} compares the estimators' MSEs (normalized by the true value $V(\pi)$) with varying data sizes. The result clearly suggests that AIPS (ours) achieves the best (lowest) MSE in a range of logged data sizes, while the existing estimators fail drastically in some specific cases. First, we observe that IIPS and RIPS fail to improve their MSE even with increasing logged data sizes. We attribute this to their high bias due to the mismatch between their behavior assumption (independence or cascade) and the true user behavior (which is diverse and context-dependent in our experiment). Second, we can see that IPS enables an unbiased estimation,\footnote{Note that the squared bias of IPS is not exactly zero even though this estimator is always theoretically unbiased. This is due to the fact that we estimate the squared bias based on the simulation results where there is some small variance.} however, it suffers from extreme variance, particularly when the data size is small. This is because IPS often applies unnecessarily large importance weights regardless of the true user behavior.

In contrast, AIPS (true) deals with the bias-variance issues of the existing estimators by leveraging adaptive importance weighting based on prior knowledge about the true user behavior (which is unavailable in practice). Specifically, Figure~\ref{fig:data_size} demonstrates that AIPS (true) is unbiased and thus performs better than IIPS and RIPS when the data size is large where AIPS (true) becomes increasingly accurate with a reduced variance while IIPS and RIPS remain highly biased. Moreover, AIPS (true) has a much lower variance than IPS by applying importance weighting to only the relevant set of actions for each given context. However, it should be noted that the variance of AIPS (true) can still be high, particularly when the data size is extremely small. In particular, AIPS (true) exhibits a worse MSE than RIPS when $n \leq 2\mathrm{k}$, which interestingly implies that naive use of true user behavior when performing importance weighting is not optimal in terms of MSE.

Our AIPS estimator performs much better than all existing estimators and even overcomes the limitations of AIPS (true) by \textit{optimizing} the user behavior model rather than merely exploiting the true model. More specifically, AIPS further improves the MSE of AIPS (true) by greatly reducing the variance at the cost of introducing only a small amount of bias.
Note here that this is achieved even though we impose some estimation error in the MSE estimation, suggesting that the subgroup optimization procedure from Section~\ref{sec:ubso} is robust to the estimation error of the MSE.\footnote{We also observe the similar results and superior behavior of AIPS with varying amounts of noise on the MSE estimate, which is reported in Appendix~\ref{app:experiment}.} These empirical results demonstrate that AIPS is able to adaptively optimize the user behavior model in a way that improves the MSE and thus enables a more reliable OPE in a range of logged data sizes particularly under diverse user behavior and even without the true knowledge of the behavior model.

\paragraph{\textbf{RQ (2): How do the estimators perform with varying lengths of ranking?}} 
Next, we compare the performance of the estimators with varying lengths of ranking ($K$) in Figure~\ref{fig:slate_size}. The overall trend and qualitative comparison are similar to the previous arguments made in RQ (1) -- AIPS (ours) works stably well across a range of settings, while the existing estimators fail for some specific values of $K$. Specifically, when $K \geq 10$, IPS and AIPS (true) produce extremely high variance due to excessive importance weights, while IIPS and RIPS produce substantial bias due to their strong assumption about user behavior. In contrast, AIPS (ours) leads to a much better bias-variance tradeoff by \textit{optimizing} the user behavior. In particular, we observe that AIPS (ours) prioritizes reducing bias when $K \leq 8$, while it puts more priority on variance reduction when $K \geq 10$, resulting in its superior performance against existing methods as well as AIPS (true) in a range of ranking sizes.

\paragraph{\textbf{RQ (3): How do the estimators perform with various user behavior distributions?}}
Figure~\ref{fig:user_behavior_variation} shows how the accuracy of the estimators changes as the user behavior distribution shifts from a simple user behavior ($\delta=0.0$) to a more realistic, complex one ($\delta=1.0$). 
First, the result demonstrates that IPS produces inaccurate OPE across various behavior distributions, as its variance is consistently high. In contrast, IIPS and RIPS are accurate when user behavior is simple ($\delta=0.0$). 
However, as user behavior gradually becomes more complex, IIPS and RIPS produce larger bias because their assumptions become increasingly incorrect. Similarly, AIPS (true) enables an accurate estimation, particularly when the user behavior is simple, but its MSE gradually becomes worse as the user behavior becomes more complex. Specifically, AIPS (true) is accurate when $\delta \leq 0.4$ due to its optimal variance, however, it suffers from extremely high variance due to large importance weights and shows substantial accuracy deterioration in the presence of complex user behaviors. Finally, we observe that AIPS (ours) consistently achieves a much more accurate estimation compared to IPS, IIPS, and RIPS across various behavior distributions, particularly under the challenging cases of complex user behaviors ($\delta \geq 0.6$). Moreover, AIPS (ours) is even better than AIPS (true) when $\delta \geq 0.6$, because AIPS (ours) optimizes the user behavior model and thus avoids the excessive variance of AIPS (true) as long as this strategic variance reduction does not introduce considerable bias. In the case of simple behaviors ($\delta \leq 0.4$), it becomes more important to reduce the bias by leveraging the true behavior model, and thus AIPS (true) performs the best in these cases. However, the results clearly demonstrate the benefit of AIPS against existing methods (IPS, IIPS, RIPS) and that of the idea of behavior model optimization in practical situations where the user behavior is highly complex.

\section{Real-World Experiment}
This section demonstrates the effectiveness of AIPS using the logged data collected on a real-world ranking system.

\paragraph{\textbf{Setup}}
To evaluate and compare the estimators in a  more practical situation, we collect some logged bandit data by running an A/B test of two (factored) ranking policies $\pi_A$ and $\pi_B$ on a real e-commerce platform whose aim is to optimize a ranking of modules (which showcase a set of products inside) to maximize the number of clicks. Our A/B test produces two sets of logged data $\calD_A$ and $\calD_B$ where $|\calD_A| = 1,979$ and $|\calD_B| = 1,954$. Note that, in this application, $\mx$ is a five-dimensional user context, $\ma$ is a ranking of modules where $|\calA|=2$, $K=6$, and $r_k$ is a binary click indicator.

To perform an OPE experiment, we regard $\pi_A$ as a logging policy and $\pi_B$ as an evaluation policy. We use $\calD_A$ to estimate the value of $\pi_B$ by estimators, while we use $\calD_B$ to approximate the ground truth value of $\pi_B$ by on-policy estimation. Then, we calculate the squared error (SE) of an estimator as $\mathrm{SE}(\hat{V}) := (\hat{V}_{\textit{on}}(\pi_B; \calD_B) - \hat{V}(\pi_B; \calD_A))^2$. We run the experiment 100 times using different bootstrapped samples of $\calD_A$ and report the cumulative distribution function (CDF) of SE of IPS, IIPS, RIPS, and AIPS, relative to that of AIPS. To evaluate the worst case performance of the estimators, we also report the conditional value at risk (CVaR) of SE, which measures the average performance of the worst $\alpha \times 100$\% trials for each estimator. AIPS uses $\mathcal{C} = \{\mc_{S},\mc_{C},\mc_{I},\mc',\mc''\}$ as the candidate set of behavior models where $\mc_{S}$, $\mc_{C}$, and $\mc_{I}$ are defined in Section~\ref{sec:synthetic_experiment}, and $\mc'$ and $\mc''$ are defined in Appendix~\ref{app:experiment}. When performing user behavior optimization of AIPS, its MSE is estimated by PAS-IF~\citep{udagawa2022policy} using only the observable logged data. Note that we cannot implement AIPS (true) in this section, since we do not know the true user behavior in the real-world dataset.

\paragraph{\textbf{Result}}
Figure~\ref{fig:real} (Left) shows the estimators' CDF of relative SE and demonstrates that AIPS performs the best in 76\% of the trials.  Moreover, in Figure~\ref{fig:real} (Right), we observe that AIPS improves the CVaR of SE more than 30\% compared to RIPS for a range of $\alpha$. These results suggest that AIPS enables a more accurate and stable OPE than previous estimators in the real-world situation.

\paragraph{\textbf{Summary of empirical findings}}
In summary, AIPS achieves far more accurate OPE than all existing estimators (IPS, IIPS, and RIPS) in both synthetic and real-world experiments via leveraging adaptive importance weighting with an optimized user behavior model. Specifically, AIPS has a much lower bias than IIPS and RIPS by identifying more appropriate behavior models that have a sufficient overlap with the true user behavior. Moreover, AIPS substantially reduces the variance of IPS by avoiding unnecessarily large importance weights. As a result, AIPS shows a superior performance particularly in realistic situations where the ranking size is large and user behavior is diverse and complex.\footnote{Appendix~\ref{app:experiment} provides additional experiment results demonstrating that AIPS is more robust to reward noise and changes in user behavior distribution compared to baseline estimators (IPS, IIPS, and RIPS).} Moreover, we observe that AIPS performs even better than AIPS (true) in many cases, implying that strategically leveraging an incorrect behavior model can lead to a better MSE. We thus conclude that AIPS enables a more effective OPE of ranking systems and that we should consider optimizing the behavior model to improve the MSE rather than being overly sensitive to its correct estimation.

\section{Related Work} \label{app:related}

\paragraph{\textbf{Off-Policy Evaluation}}
OPE is of great practical relevance in search and recommender systems, as it enables the performance evaluation of counterfactual policies without interacting with the actual users~\citep{saito2021counterfactual, saito2021evaluating, kiyohara2021accelerating, levine2020offline}. 
In particular, OPE in the single action setting has been studied extensively, producing many estimators with good theoretical guarantees~\citep{li2011unbiased, gilotte2018offline, kallus2020optimal, saito2022off}.
Among them, IPS is often considered a benchmark estimator~\citep{precup2000eligibility}, which uses the importance sampling technique to correct the distribution shift between different policies. IPS is unbiased under some identification assumptions such as full support and unconfoundedness, but it often suffers from high variance~\citep{dudik2011doubly}. Doubly Robust (DR)~\citep{dudik2011doubly} reduces the variance of IPS by using an estimated reward function as a control variate. However, DR can still struggle with high variance when the action space is extremely large~\citep{saito2022off} such as in the ranking setup. 

Beyond the standard OPE, there has also been a growing interest in OPE of ranking systems due to its much practical relevance. In the ranking setting where a policy chooses a ranked list of items to present to the users, OPE faces the critical variance issue due to combinatorial action spaces. To address this variance issue, existing work has introduced some assumptions about user behavior. In particular, IIPS~\citep{li2018offline} assumes that a user interacts with the actions independently across positions. Under this assumption, the reward observed at each position depends only on the action presented at the same position, leading to a significant variance reduction compared to IPS. Although IIPS is unbiased when the independence assumption holds true, it can have a large bias when users follow a more complicated behavior~\citep{mcinerney2020counterfactual, kiyohara2022doubly}. RIPS~\citep{mcinerney2020counterfactual} assumes a more reasonable assumption, called the cascade assumption, which requires that a user interacts with the actions sequentially from top to bottom~\citep{guo2009efficient}. Therefore, the reward observed at each position depends only on the actions presented at higher positions. Leveraging the cascade assumption, RIPS can somewhat reduce the variance of IPS while being unbiased in more realistic cases compared to IIPS. To further improve the variance of RIPS, \citet{kiyohara2022doubly} propose the Cascade-DR estimator, leveraging the recursive structure of the cascade assumption and a control variate. Although the above approach has shown some empirical success, the critical issue is that all the above estimators rely on a single assumption (independence or cascade) applied to every user, which can cause large bias and unnecessary variance. Therefore, we were based on a more general formulation by assuming that user behavior is sampled from some unknown context-dependent distribution. As a result, AIPS provides an unbiased estimation even under arbitrarily diverse user behavior and achieves the minimum variance among the class of IPS estimators that are unbiased. Moreover, we developed a method to \textit{optimize} the user behavior model rather than accurately estimating it given the theoretical observations that the true user behavior is not optimal in terms of MSE.

Note that there is another estimator called the Pseudo Inverse (PI) estimator~\citep{swaminathan2017off, vlassis2021off,su2020doubly} in the slate recommendation setting. This estimator considers a situation where only the slate-wise reward is observed (i.e., the position-wise rewards are unobservable). Since PI is not able to leverage position-wise rewards, it is often highly sub-optimal in our setup where position-wise rewards are observable, as empirically verified in~\citet{mcinerney2020counterfactual}.

\paragraph{\textbf{Click Models}}

The click models aim to formulate how users interact with a list of documents~\citep{dupret2008user, guo2009efficient, chapelle2009dynamic, chuklin2015click, wang2016learning, joachims2017unbiased, saito2020unbiased, saito2020pairwise, saito2020dual}, and it has typically been studied based on the following \textit{examination hypothesis}: $p(c_k = 1 \,|\,\mx, \ma)= p(o_k = 1 \,|\,\mx,\ma) \cdot p(r_k=1 \,|\,\mx, a_k)$, where $c_k$ is a click indicator while $r_k$ is a relevance indicator of the document presented at the $k$-th position. $p(o_k \,|\, \mx, \ma)$ is the probability that a user examines the $k$-th document in a ranking. When the user examines the document (i.e., $o_k = 1$), the click probability is assumed identical to the probability of relevance. Much research has been done to better parameterize the examination probability to explain finer details of the real-world examination behavior. For example, the Position-based model assumes that the examination probability depends only on the position in a ranking, while the Cascade model~\citep{craswell2008experimental,guo2009efficient} assumes that the examination probability at the $k$-th position depends on the relevance of the documents shown at higher positions.

In contrast, the user behavior models utilized in OPE focus more on modeling the dependencies among actions and rewards rather than modeling the examination probability~\citep{mcinerney2020counterfactual,kiyohara2022doubly}. As already discussed, the critical drawback of the previous methods is that only a single assumption is assumed to model every user's behavior. In the information retrieval literature, \citet{chen2020context} considered context-dependent click models, which assume that the examination behavior may change depending on the search query. Moreover, several studies have indicated the need to incorporate some context information in building and estimating click models such as devices~\citep{mao2018constructing}, user browsing history~\citep{deng2020hybrid}, and user intention~\citep{hu2011characterizing}. In this work, we deal with potentially diverse user behaviors by formulating them via a context-dependent distribution for the first time in OPE of ranking policies. Note, however, that our motivation is substantially different from that of the click modeling literature. That is, we aim to develop an accurate OPE estimator in terms of MSE while click modeling aims to estimate the true user behavior as accurately as possible. This difference motivates our unique strategy to intentionally rely on an incorrect behavior model to further improve the MSE of our estimator as discussed in Section~\ref{sec:ubso}.

\section{Conclusion and Future Work}
This paper studied OPE of ranking systems under diverse user behavior. When the user behavior is diverse and depends on the user context, all existing estimators can be highly sub-optimal because they apply a single assumption to the entire population. To achieve an effective OPE even under much more diverse user behavior, we propose the \textit{Adaptive IPS} estimator based on a new formulation where the user behavior is assumed to be sampled from a \textit{context-dependent} distribution. We began by theoretically characterizing the bias and variance of AIPS assuming known user behaviors, showing that it can be unbiased under any distribution of user behavior and that it achieves the optimal variance among unbiased IPS estimators. Interestingly, though, our analysis also indicates that myopically using the true user behavior in OPE might not be optimal in terms of MSE. Therefore, we provided a data-driven procedure to \textit{optimize} the user behavior model to minimize the MSE of the resulting AIPS estimator rather than trying to \textit{estimate} the true behavior, which tends to be sub-optimal in OPE. Experiments demonstrate that AIPS provides a substantial gain in MSE against existing methods in a range of OPE situations.

Our work also raises several intriguing research questions for future studies. First, it would be valuable to develop an accurate way to estimate the MSE of an OPE estimator beyond existing methods~\citep{su2020doubly,udagawa2022policy} to better optimize the user behavior model to further improve AIPS. Second, OPE of ranking policies can still become extremely difficult when the number of unique actions ($|\calA|$) is large. Therefore, it would be interesting to leverage the recent action embedding approach~\citep{saito2022off,saito2023off} to overcome this critical limitation in the ranking setup. Besides, as a practical, yet simple extension, adding a control variate to AIPS is expected to further improve its variance and outperform Cascade-DR of~\citet{kiyohara2022doubly}, which assumes the cascade assumption. Finally, this work only studied the statistical problem of estimating the value of a fixed new policy, so it would be interesting to use our estimator to enable more efficient off-policy learning in ranking systems.

%% file: appendix.tex
\appendix
\clearpage
\onecolumn

\begin{figure*}[t]
\begin{minipage}[c]{0.99\hsize}
\centering
\scalebox{0.95}{
\begin{tabular}{c}
\begin{minipage}{0.50\hsize}
\begin{center}
\includegraphics[width=1.0\linewidth]{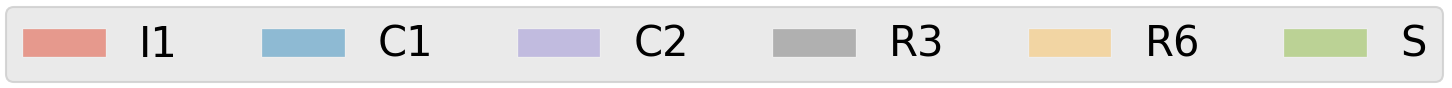}
\end{center}
\end{minipage}
\\
\\
\begin{minipage}{0.99\hsize}
    \begin{center}
        \includegraphics[clip, width=1.0\linewidth]{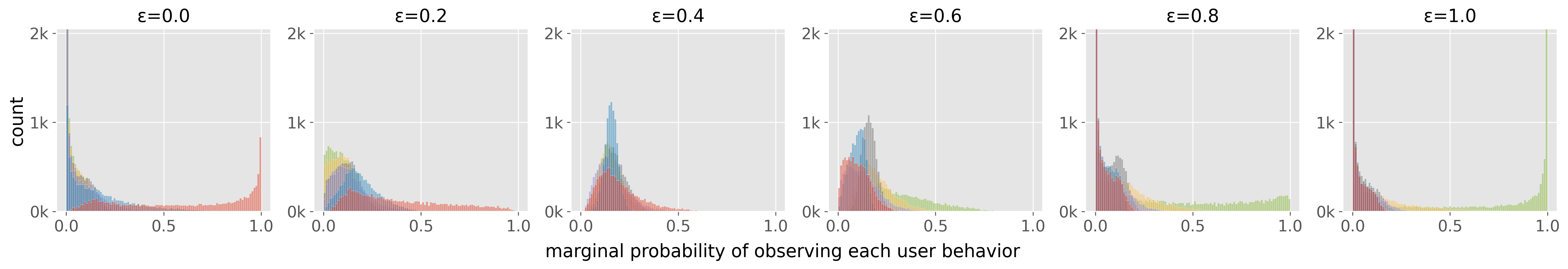}
        \vspace{-6mm}
        \caption{User behavior distribution with varying values of $\delta$}
        \label{fig:variation_setting}
        \vspace{5mm}
    \end{center}
\end{minipage}
\\ 
\\
\begin{minipage}{0.99\hsize}
    \begin{center}
        \includegraphics[clip, width=1.0\linewidth]{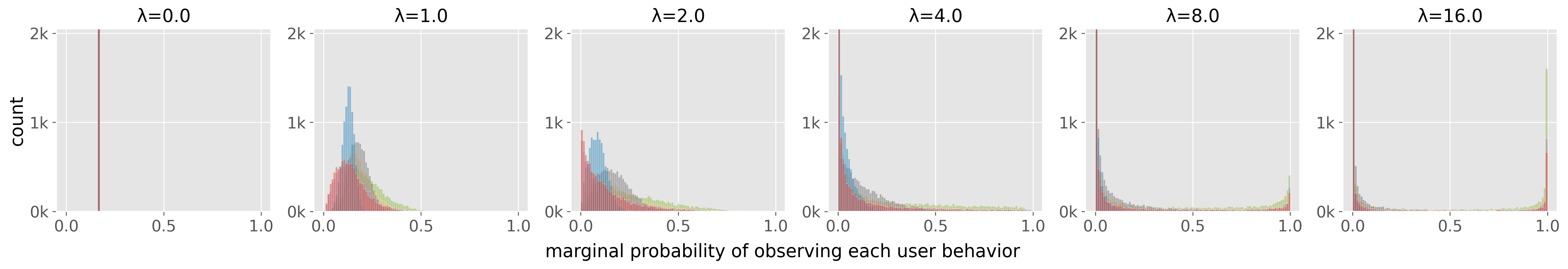}
        \vspace{-6mm}
        \caption{User behavior distribution with varying values of $\lambda$}
        \label{fig:stochasticity_setting}
        \vspace{5mm}
    \end{center}
\end{minipage}
\\
\\
\end{tabular}
}
\end{minipage}
\end{figure*}

\begin{figure}[tb]
    \centering
    \includegraphics[clip, width=6.0cm]{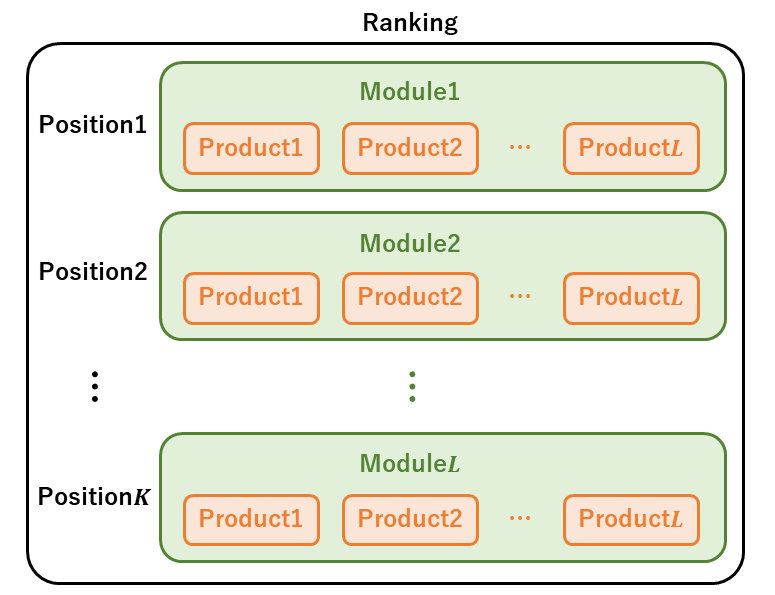}
    \caption{This figure illustrates a ranking of modules in the e-commerce platform used in our real-world experiment where each "Module" corresponds to an action indicating a category of products, such as "Recommended items" or "Campaign information".}
    \label{fig:module}
\end{figure}

\section{Additional Experiment Details and Results} \label{app:experiment}

\subsection{Experimental Details}
\paragraph{\textbf{Distributions of user behavior in the synthetic experiment}} In the synthetic experiment, we control the distribution of user behavior by varying the values of $\delta$ (user behavior distribution parameter) -- a small value of $\delta$ increases the probability of observing simple user behaviors, while a large value of $\delta$ increases the probability of observing complex user behaviors. Figure~\ref{fig:variation_setting} demonstrates how different values of $\delta$ control the distribution of user behavior, which we estimate with randomly sampled 10,000 user contexts.

\paragraph{\textbf{Platform's ranking interface in the real-world experiment}} 
Figure~\ref{fig:module} illustrates the ranking interface of the e-commerce platform used in the real-world experiment. The two factored policies, $\pi_A$ and $\pi_B$, choose which module as an action to present at each position in a ranking to maximize the sum of observed clicks during the data collection experiment.

\paragraph{\textbf{The candidate set of behavior models for AIPS in the real-world experiment}}
In the real-world experiment, AIPS uses $\mathcal{C} = \{\mc_{S},\mc_{C},\mc_{I},\mc',\mc''\}$ as the candidate set of behavior models when performing user behavior optimization. $\mc_{S}$, $\mc_{C}$, and $\mc_{I}$ are defined in Section~\ref{sec:synthetic_experiment}. $\mc'$ and $\mc''$ are defined specifically as
\begin{itemize}
    \item $\mc'$: $\mc'(k, 1)=\mc'(k, 2)=\mc'(k, k)=1$, otherwise, $\mc'(k,l) = 0, \quad \forall l \in [K]$
    \item $\mc''$: $\mc''(k, l)=1$ if $l \le k$, otherwise, $\mc''(k,l) = 0, \quad \forall l \in [K]$
\end{itemize}
for each position $k \in [K]$.

\begin{figure*}[t]
\begin{minipage}[c]{0.99\hsize}
\centering
\scalebox{0.95}{
\begin{tabular}{c}
\begin{minipage}{0.60\hsize}
\begin{center}
\includegraphics[width=1.0\linewidth]{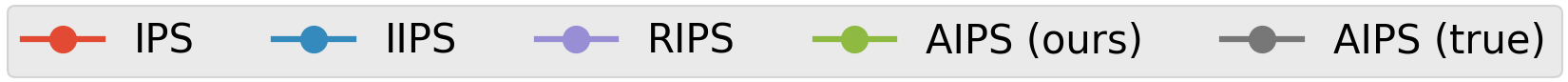}
\end{center}
\end{minipage}
\\
\\
\begin{minipage}{0.99\hsize}
    \begin{center}
        \includegraphics[clip, width=1.0\linewidth]{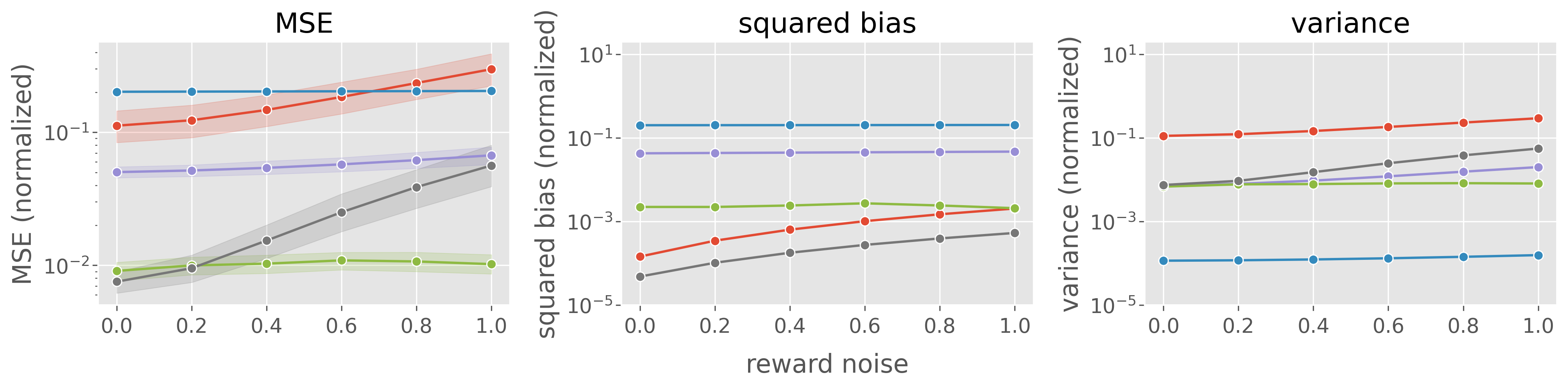}
        \vspace{-6mm}
        \caption{Comparison of the estimators' MSE (normalized by the true value $V(\pi)$) with varying reward noise levels ($\sigma$)}
        \label{fig:reward_noise}
        \vspace{5mm}
    \end{center}
\end{minipage}
\\ 
\\
\begin{minipage}{0.99\hsize}
    \begin{center}
        \includegraphics[clip, width=1.0\linewidth]{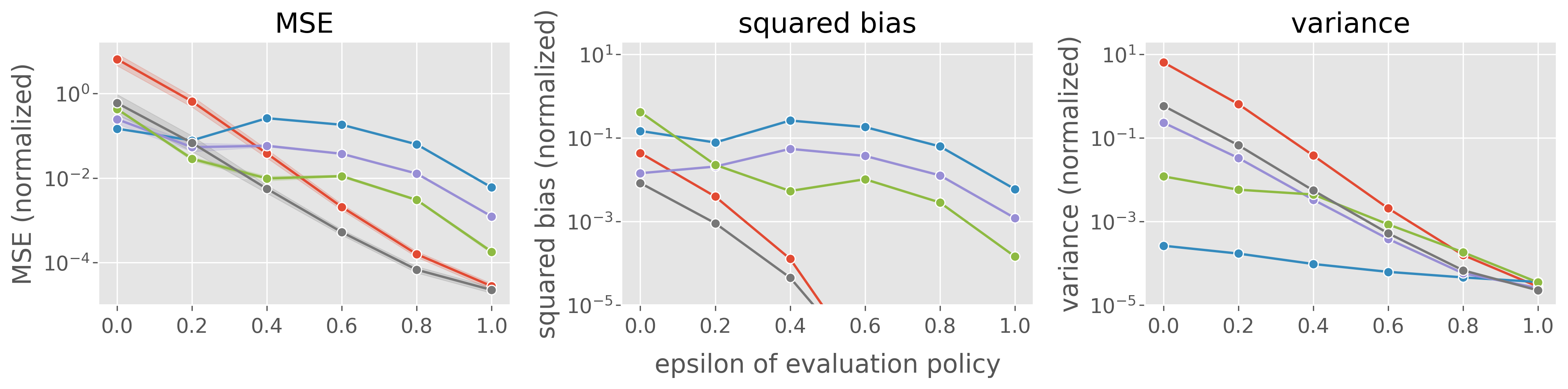}
        \vspace{-6mm}
        \caption{Comparison of the estimators' MSE (normalized by the true value $V(\pi)$) with varying evaluation policies ($\epsilon$)}
        \label{fig:evaluation_policy}
        \vspace{5mm}
    \end{center}
\end{minipage}
\\
\\
\begin{minipage}{0.99\hsize}
    \begin{center}
        \includegraphics[clip, width=1.0\linewidth]{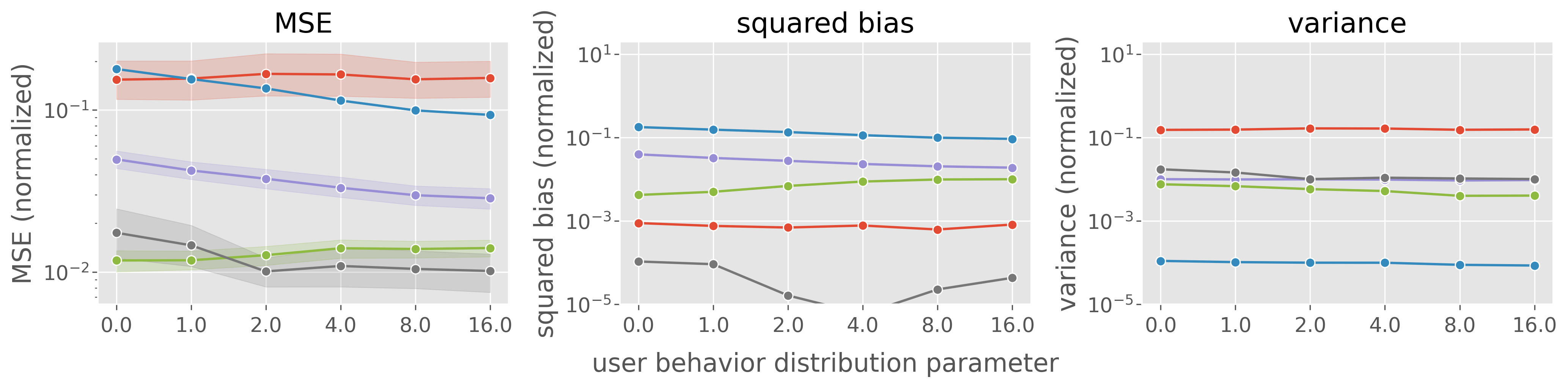}
        \vspace{-6mm}
        \caption{Comparison of the estimators' MSE (normalized by the true value $V(\pi)$) with varying behavior distributions ($\lambda$)}
        \label{fig:user_behavior_stochasticity}
    \end{center}
\end{minipage}
\\
\\
\end{tabular}
}
\end{minipage}
\end{figure*}

\begin{figure*}[!htb]
\begin{minipage}[c]{0.99\hsize}
\centering
\scalebox{0.95}{
\begin{tabular}{c}
\begin{minipage}{1.00\hsize}
\begin{center}
\includegraphics[width=1.0\linewidth]{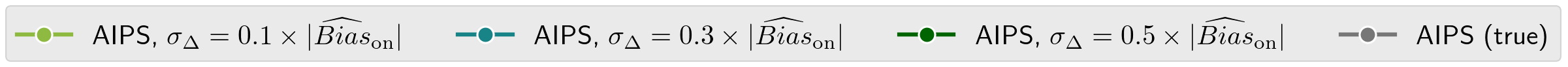}
\end{center}
\end{minipage}
\\
\\
\begin{minipage}{0.99\hsize}
    \begin{center}
        \includegraphics[clip, width=1.0\linewidth]{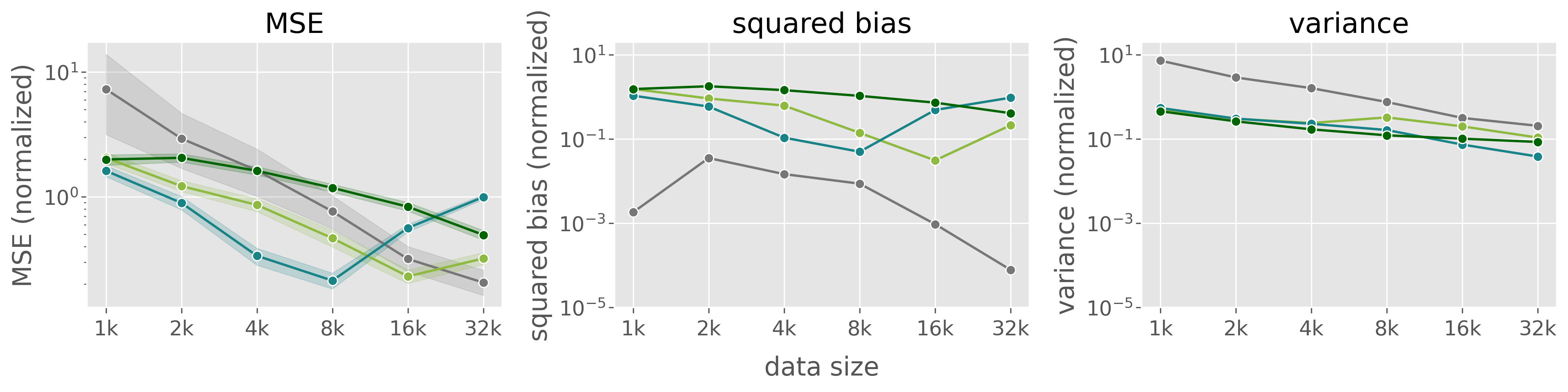}
        \vspace{-6mm}
        \caption{Comparison of AIPS's performance with varying data sizes ($n$) and estimation error ($\sigma_{\Delta}$)}
        \label{fig:data_size_err}
        \vspace{5mm}
    \end{center}
\end{minipage}
\\ 
\\
\begin{minipage}{0.99\hsize}
    \begin{center}
        \includegraphics[clip, width=1.0\linewidth]{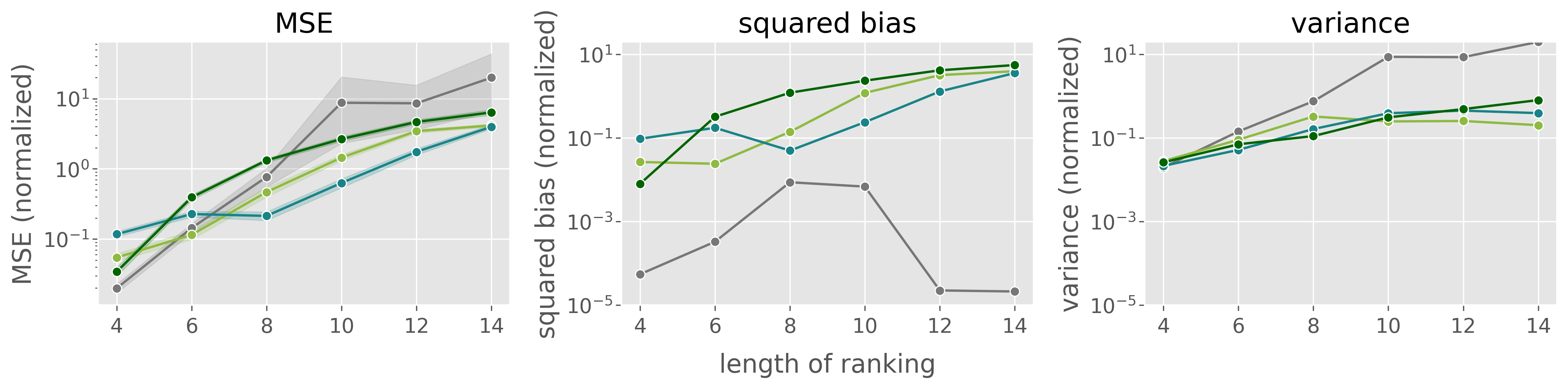}
        \vspace{-6mm}
        \caption{Comparison of AIPS's performance with varying lengths of ranking ($K$) and estimation error ($\sigma_{\Delta}$)}
        \label{fig:slate_size_err}
        \vspace{5mm}
    \end{center}
\end{minipage}
\\
\\
\begin{minipage}{0.99\hsize}
    \begin{center}
        \includegraphics[clip, width=1.0\linewidth]{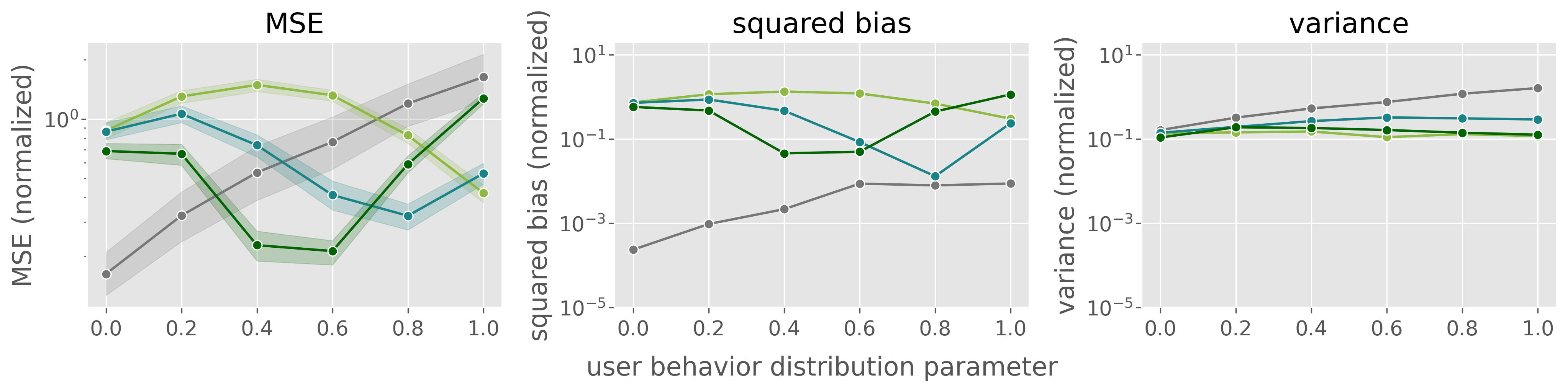}
        \vspace{-6mm}
        \caption{Comparison of AIPS's performance with varying user behavior distributions ($\delta$) and estimation error ($\sigma_{\Delta}$)}
        \label{fig:user_behavior_variation_err}
    \end{center}
\end{minipage}
\\
\\
\end{tabular}
}
\end{minipage}
\end{figure*}

\begin{figure*}[!htb]
\begin{minipage}[c]{0.99\hsize}
\centering
\scalebox{0.95}{
\begin{tabular}{c}
\begin{minipage}{0.75\hsize}
\begin{center}
\includegraphics[width=1.0\linewidth]{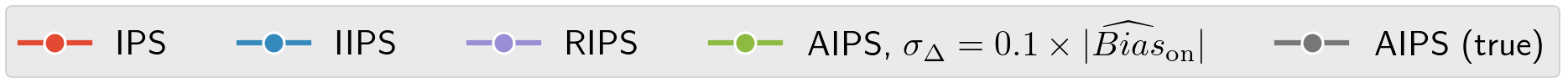}
\end{center}
\end{minipage}
\\
\\
\begin{minipage}{0.99\hsize}
    \begin{center}
        \includegraphics[clip, width=1.0\linewidth]{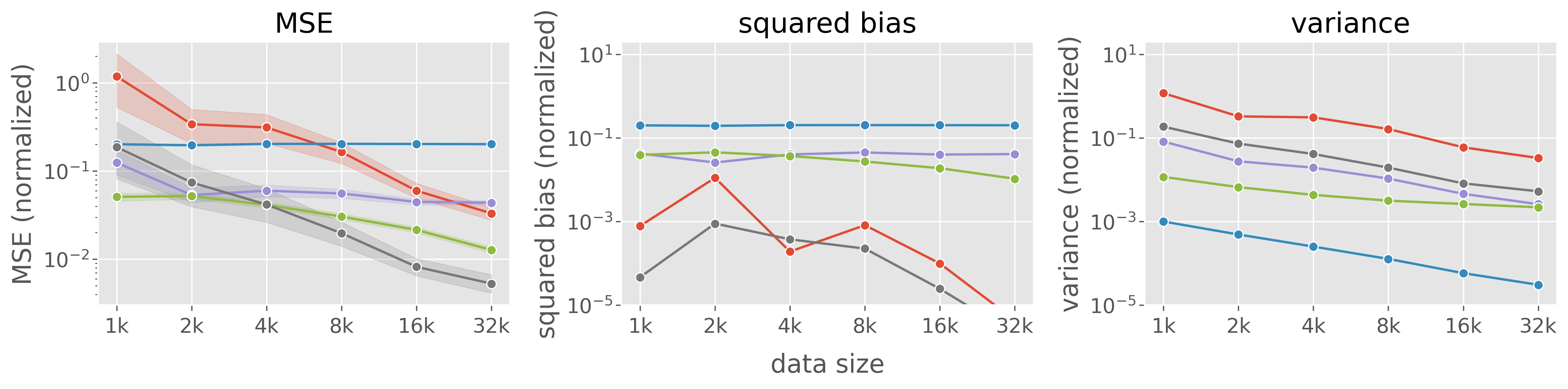}
        \vspace{-6mm}
        \caption{Comparison of the estimators' MSE (normalized by the true value $V(\pi)$) with varying data sizes ($n$) when $\sigma_{\Delta} = 0.1 \times |\widehat{Bias}_{\mathrm{on}}|$}
        \label{fig:data_size_err_small}
        \vspace{5mm}
    \end{center}
\end{minipage}
\\ 
\\
\begin{minipage}{0.99\hsize}
    \begin{center}
        \includegraphics[clip, width=1.0\linewidth]{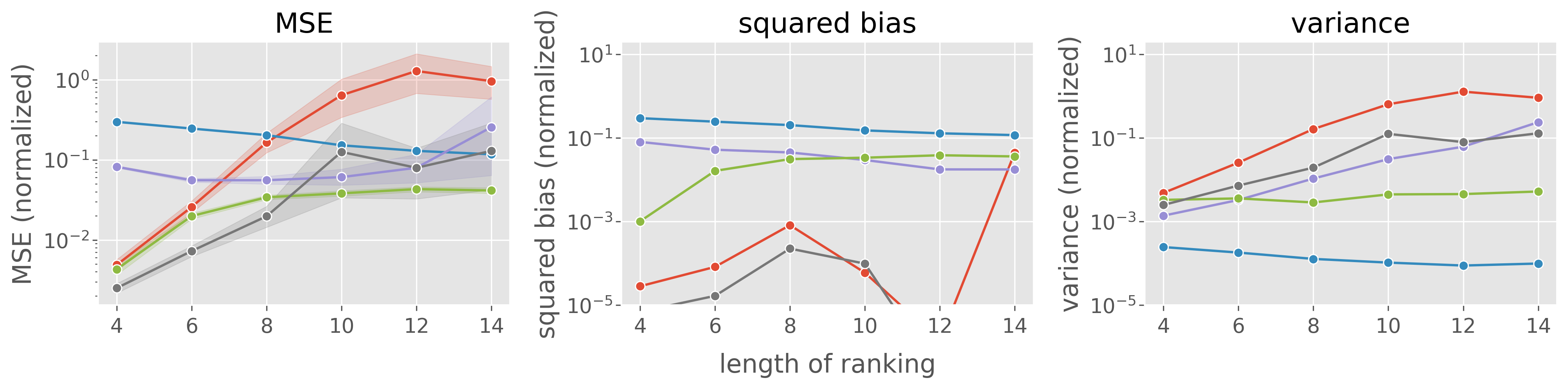}
        \vspace{-6mm}
        \caption{Comparison of the estimators' MSE (normalized by the true value $V(\pi)$) with varying lengths of ranking ($K$) when $\sigma_{\Delta} = 0.1 \times |\widehat{Bias}_{\mathrm{on}}|$}
        \label{fig:slate_size_err_small}
        \vspace{5mm}
    \end{center}
\end{minipage}
\\
\\
\begin{minipage}{0.99\hsize}
    \begin{center}
        \includegraphics[clip, width=1.0\linewidth]{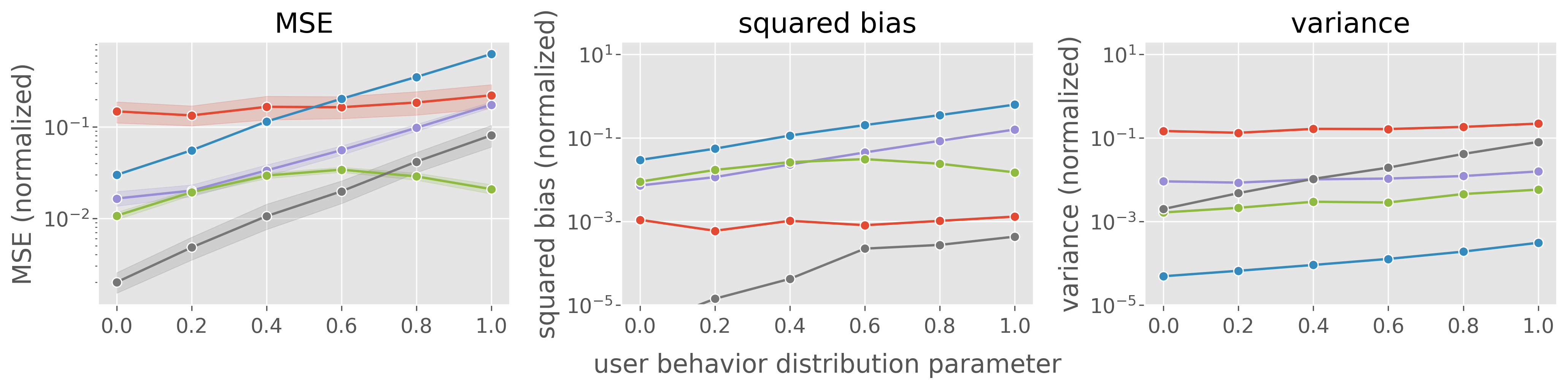}
        \vspace{-6mm}
        \caption{Comparison of the estimators' MSE (normalized by the true value $V(\pi)$) with varying user behavior distributions ($\delta$) when $\sigma_{\Delta} = 0.1 \times |\widehat{Bias}_{\mathrm{on}}|$}
        \label{fig:user_behavior_variation_err_small}
    \end{center}
\end{minipage}
\\
\\
\end{tabular}
}
\end{minipage}
\end{figure*}

\begin{figure*}[!htb]
\begin{minipage}[c]{0.99\hsize}
\centering
\scalebox{0.95}{
\begin{tabular}{c}
\begin{minipage}{0.75\hsize}
\begin{center}
\includegraphics[width=1.0\linewidth]{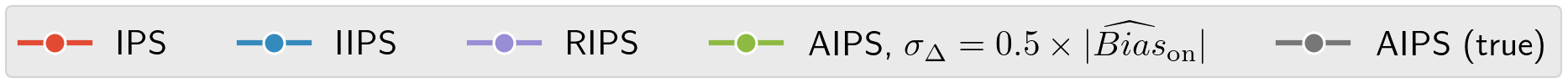}
\end{center}
\end{minipage}
\\
\\
\begin{minipage}{0.99\hsize}
    \begin{center}
        \includegraphics[clip, width=1.0\linewidth]{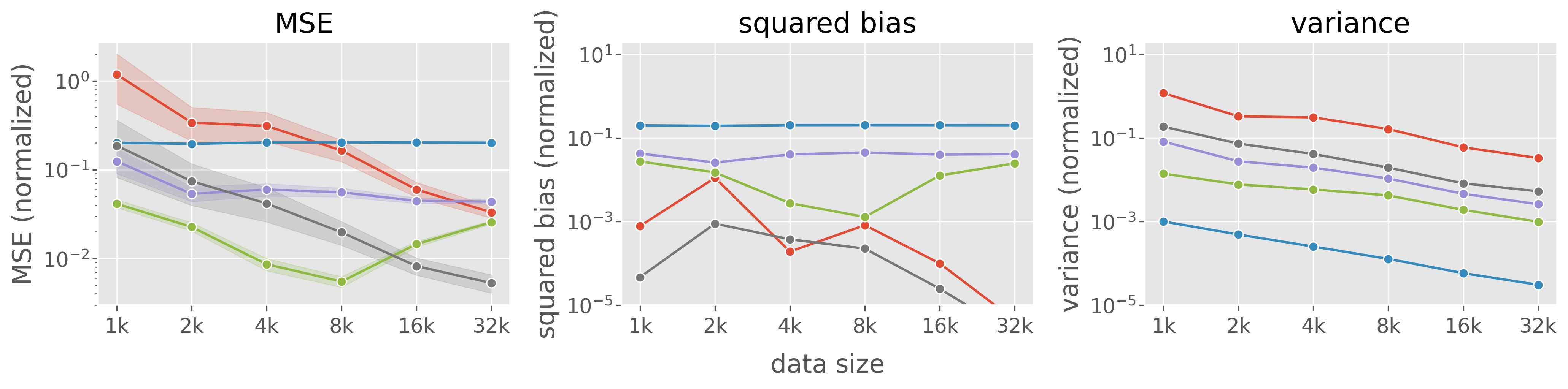}
        \vspace{-6mm}
        \caption{Comparison of the estimators' MSE (normalized by the true value $V(\pi)$) with varying data sizes ($n$) and $\sigma_{\Delta} = 0.5 \times |\widehat{Bias}_{\mathrm{on}}|$}
        \label{fig:data_size_err_karge}
        \vspace{5mm}
    \end{center}
\end{minipage}
\\ 
\\
\begin{minipage}{0.99\hsize}
    \begin{center}
        \includegraphics[clip, width=1.0\linewidth]{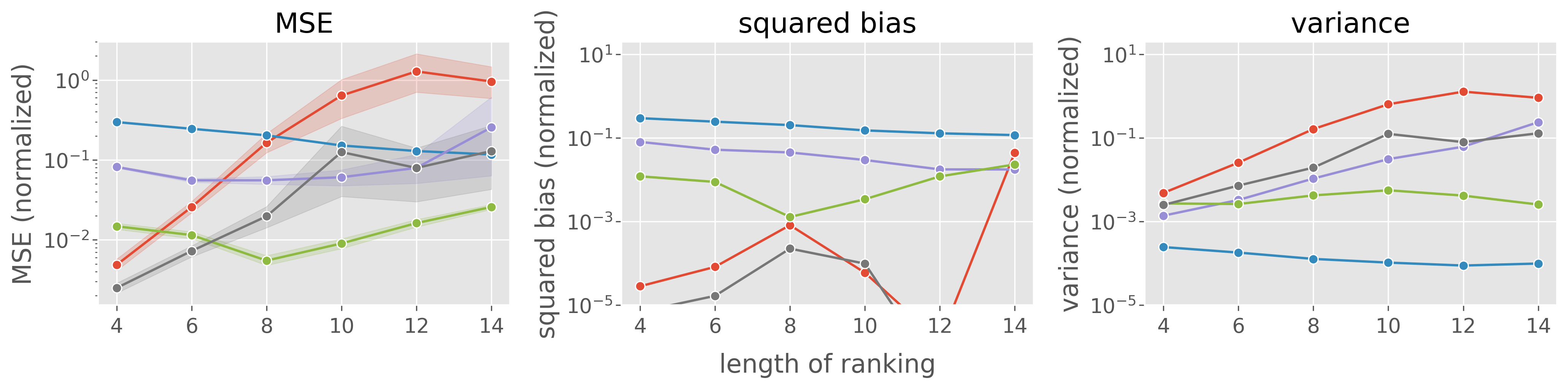}
        \vspace{-6mm}
        \caption{Comparison of the estimators' MSE (normalized by the true value $V(\pi)$) with varying lengths of ranking ($K$) and $\sigma_{\Delta} = 0.5 \times |\widehat{Bias}_{\mathrm{on}}|$}
        \label{fig:slate_size_err_karge}
        \vspace{5mm}
    \end{center}
\end{minipage}
\\
\\
\begin{minipage}{0.99\hsize}
    \begin{center}
        \includegraphics[clip, width=1.0\linewidth]{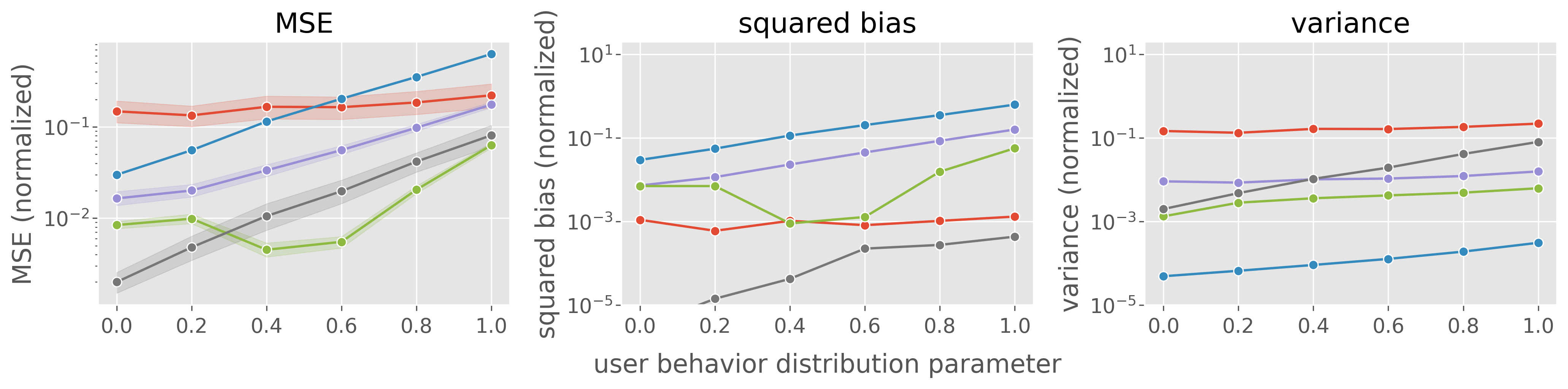}
        \vspace{-6mm}
        \caption{Comparison of the estimators' MSE (normalized by the true value $V(\pi)$) with varying user behavior distributions ($\delta$) and $\sigma_{\Delta} = 0.5 \times |\widehat{Bias}_{\mathrm{on}}|$}
        \label{fig:user_behavior_variation_err_karge}
    \end{center}
\end{minipage}
\\
\\
\end{tabular}
}
\end{minipage}
\end{figure*}

\subsection{Additional Results on Synthetic Data}
This section explores four additional research questions regarding: (A1) reward noise level ($\sigma$), (A2) evaluation policy ($\epsilon$), (A3) identifiability of user behavior ($\lambda$), and (A4) estimation error in the bias estimate ($\sigma_{\Delta}$) used in AIPS. Note that we set $n=8\mathrm{k}$ (data size), $K=8$ (length of ranking), $\sigma=0.5$ (reward noise level), $\epsilon=0.3$ (evaluation policy parameter), $\delta=0.6$ (user behavior setting), and $\sigma_{\Delta}=0.3$ (error level in bias estimate) as default, and run OPE simulations over 300 different logged data replicated with different random seeds. The following reports and discusses the MSE, squared bias, and variance of the estimators normalized by the true policy value of the evaluation policy $V(\pi)$.

\paragraph{\textbf{RQ (A1): How do the estimators perform with varying reward noise levels?}} 
Figure~\ref{fig:reward_noise} compares estimators' performance with varying reward noise levels $\sigma \in \{0.0, 0.2, \cdots, 1.0 \}.$ We observe that IPS and AIPS (true) suffer from increasingly high variance as the noise level becomes higher. In contrast, our AIPS demonstrates its clear robustness to the increase in reward noise.

\paragraph{\textbf{RQ (A2): How do the estimators perform with varying evaluation policies?}} 
Next, Figure~\ref{fig:evaluation_policy} compares the estimators with various evaluation policies $\epsilon \in \{0.0, 0.2, \cdots, 1.0\}$. The figure indicates that AIPS (true) and IPS are quite accurate when the evaluation policy is near-uniform and does not deviate from the logging policy greatly ($\epsilon = 0.8, 1.0$). This is because AIPS (true) and IPS are unbiased and do not suffer from high variance when the evaluation policy is highly stochastic. However, under more practical situations where the evaluation policy is more deterministic, AIPS becomes superior due to its favorable variance property while IPS performs the worst due to its extreme variance. Note that AIPS (true) also works well for a range of evaluation policies in the default setting (i.e., $n=8\mathrm{k}, K=8$). However, it may suffer from a higher variance when the ranking size is larger ($K > 8$) and the evaluation policy is more deterministic ($\epsilon=0.0, 0.2$), as already shown in Figure~\ref{fig:slate_size}.

\paragraph{\textbf{RQ (A3): How do the estimators perform under different levels of identifiability of user behavior?}} 
Here, we investigate how the estimators perform with varying levels of identifiability of user behavior. Specifically, we control the identifiability of user behavior by varying the values of $\lambda_z$. Recall here that we sample user behavior from the following conditional distribution.
\begin{align*}
    p(\mc_z\,|\,\mx) 
    := \operatorname{softmax} (\lambda_z \cdot |\theta_z^{\top} \mx|)
    = \frac{\exp(\lambda_z \cdot |\theta_z^{\top} \mx|)}{\sum_{z'} \exp(\lambda_{z'} \cdot |\theta_{z'}^{\top} \mx|)}.
\end{align*}
Thus, by definition, when we set $\lambda_z = \lambda, \forall z$ (constant value), user behavior will be uniformally distributed. We can increase identifiability of user behavior by using a large value of $\lambda$ where user behavior becomes near-deterministic and easily identifiable from the context. By contrast, when we use a small value of $\lambda$, user behavior will be almost context-independent and may not be easily identifiable from the observed user context. We thus vary $\lambda \in \{ 0.0, 1.0, 2.0, \cdots, 16.0 \}$ to see how the estimators' performance changes with different levels of identifiability of user behavior (Figure~\ref{fig:stochasticity_setting} illustrates the distributions of user behavior with varying values of $\lambda$).

Somewhat surprisingly, we observe in Figure~\ref{fig:user_behavior_stochasticity} that identifiability of user behavior has almost no impact on the estimators' performances. We conjuncture that AIPS is robust to the change in the level of identifiability because it does not aim to precisely estimate user behavior, but it rather aims to choose the most suitable user behavior model to minimize its MSE given the logged data. This observation provides a further argument for the applicability and robustness of the user behavior optimization procedure of AIPS.

\paragraph{\textbf{RQ (A4): How does AIPS perform with varying estimation errors in the bias estimate?}}
Finally, we evaluate how robust AIPS is to the estimation error in the bias estimate by varying the values of  $\sigma_{\Delta} \in \{ 
0.1, 0.3, 0.5 \}$ $\times |\widehat{Bias}_{\mathrm{on}}|$. More specifically, we compare AIPS' performance with the three different values of $\sigma_{\Delta}$ across varying data sizes $n \in \{1\mathrm{k}, 2\mathrm{k}, \cdots, 32\mathrm{k} \}$ (Figure~\ref{fig:data_size_err}), lengths of ranking $K \in \{4, 6, \cdots, 14\}$ (Figure~\ref{fig:slate_size_err}), and user behavior distribution parameters $\delta \in \{ 0.0, 0.2, \cdots, 1.0 \}$ (Figure~\ref{fig:user_behavior_variation_err}) as done in RQs (1)-(3) in the main text. We also compare AIPS against IPS, IIPS, and RIPS with $\sigma_{\Delta} = 0.1 \times |\widehat{Bias}_{\mathrm{on}}|$ in Figures~\ref{fig:data_size_err_small}-\ref{fig:user_behavior_variation_err_small} and with $\sigma_{\Delta} = 0.5 \times |\widehat{Bias}_{\mathrm{on}}|$ in Figures~\ref{fig:data_size_err_karge}-\ref{fig:user_behavior_variation_err_karge}.

Overall, the results indicate that AIPS is robust to the estimation error in the bias estimate. In particular, the trends observed in Figures~\ref{fig:data_size_err_small}-\ref{fig:user_behavior_variation_err_small} and Figures~\ref{fig:data_size_err_karge}-\ref{fig:user_behavior_variation_err_karge} are quite similar to those observed in Figures~\ref{fig:data_size}-\ref{fig:user_behavior_variation} in the main text, suggesting that AIPS effectively balances the bias-variance tradeoff even in the presence of severe estimation error in the bias estimate. Interestingly, we also observe that a larger estimation error in the bias estimate does not necessarily lead to a larger MSE. Specifically, in Figure~\ref{fig:data_size_err} and Figure~\ref{fig:slate_size_err}, the accuracy of AIPS only becomes worse with larger estimation error ($\sigma_{\Delta}$) when the data size is large ($n = 16\mathrm{k}, 32\mathrm{k}$) and the ranking size is small ($K=4,6$). This implies that the estimation error of the bias estimate may be slightly problematic only when the bias is dominant in the MSE.

\begin{table*}[t]
\large
\centering
\caption{Runtime comparison between AIPS and IPS with varying data sizes ($n$)} \label{tab:behavior}
\vspace{-3mm}
\def\arraystretch{1.2}
\scalebox{0.8}{
\begin{tabular}{c|ccccccc}
\toprule
data size ($n$) & 1,000 & 2,000 & 4,000 & 8,000 & 16,000 & 32,000 & 
\\\midrule \midrule
IPS & 0.6022 ($\pm$ 0.008) & 1.198 ($\pm$ 0.021) & 2.379 ($\pm$ 0.041) & 4.752 ($\pm$ 0.072) & 9.505 ($\pm$ 0.149) & 18.98 ($\pm$ 0.251) \\
AIPS & 23.43 ($\pm$ 0.534) & 44.22 ($\pm$ 0.9206) & 85.24 ($\pm$ 1.502) & 167.2 ($\pm$ 2.556) & 329.6 ($\pm$ 4.162)	& 653.5 ($\pm$ 7.611) \\
\midrule
(relative) & 38.90 & 36.98 & 35.83 & 35.18 & 34.68 & 34.42 \\
\bottomrule
\end{tabular}
}
\vskip 0.1in
\raggedright
\fontsize{8.5pt}{8.5pt}\selectfont \textit{Note}:
We report mean ($\pm$ std) of the runtime (sec) of IPS and AIPS in the synthetic experiment over 100 random seeds. (relative) reports the runtime of AIPS divided by that of IPS.
\label{tab:runtime_data_size}
\end{table*}

\begin{table*}[t]
\large
\centering
\caption{Runtime comparison between AIPS and IPS with varying lengths of ranking ($K$)} \label{tab:behavior}
\vspace{-3mm}
\def\arraystretch{1.2}
\scalebox{0.8}{
\begin{tabular}{c|ccccccc}
\toprule
length of ranking ($K$) & 4 & 6 & 8 & 10 & 12 & 14 & 
\\\midrule \midrule
IPS & 2.313 ($\pm$ 0.021) & 3.390 ($\pm$ 0.025) & 4.690 ($\pm$ 0.037) & 6.192 ($\pm$ 0.039) & 7.924 ($\pm$ 0.066) & 9.878 ($\pm$ 0.068)
 \\
AIPS & 68.89 ($\pm$ 0.977) & 114.3 ($\pm$ 1.212) & 166.2 ($\pm$ 1.822) & 221.5 ($\pm$ 2.384) & 284.2 ($\pm$ 3.136) & 351.5 ($\pm$ 3.059) \\
\midrule
(relative) & 29.78 & 33.70 & 35.44 & 35.76 & 35.87 & 35.59 \\
\bottomrule
\end{tabular}
}
\vskip 0.1in
\raggedright
\fontsize{8.5pt}{8.5pt}\selectfont \textit{Note}:
We report mean ($\pm$ std) of the runtime (sec) of IPS and AIPS in the synthetic experiment over 100 random seeds. (relative) reports the runtime of AIPS divided by that of IPS.
\label{tab:runtime_slate_size}
\end{table*}

\paragraph{Runtime analysis}
One potential concern of AIPS is the additional computation overhead introduced by its user behavior optimization procedure. Tables~\ref{tab:runtime_data_size} and \ref{tab:runtime_slate_size} compare the computation time of AIPS against that of IPS with varying data sizes ($n$) and lengths of ranking ($K$). The result shows that the whole estimation process ends only in 653 seconds (< 11mins) even in the largest sample size. Moreover, we can see that the relative computation time of AIPS compared to IPS does not grow with the sample size and lengths of ranking.

\section{Omitted Proofs} \label{app:proof}

\subsection{Proof of Proposition~\ref{prop:unbiased}}
\begin{proof}
For any given $\mx \sim p(\mx)$ and $\mc \sim p(\mc\,|\,\mx)$, we have
\begin{align}
    \mE_{\pi_0(\ma|\mx)p(\mr|\mx,\ma,\mc)} \left[ \frac{\pi(\Phi_k(\ma, \mc)\,|\,\mx)}{\pi_0(\Phi_k(\ma, \mc)\,|\,\mx)} r_k \right] 
    &= \sum_{\ma} \pi_0(\ma\,|\,\mx) \frac{\pi(\Phi_k(\ma,\mc)\,|\,\mx)}{\pi_0(\Phi_k(\ma,\mc)\,|\,\mx)} q_k(\mx,\ma,\mc) \notag \\
    &= \sum_{\Phi_k(\ma,\mc)} \sum_{\Phi_k^c(\ma,\mc)} \pi_0(\ma\,|\,\mx) \frac{\pi(\Phi_k(\ma,\mc)\,|\,\mx)}{\pi_0(\Phi_k(\ma,\mc)\,|\,\mx)} q_k(\mx,\ma,\mc) \notag \\
    &= \sum_{\Phi_k(\ma,\mc)} \frac{\pi(\Phi_k(\ma,\mc)\,|\,\mx)}{\pi_0(\Phi_k(\ma,\mc)\,|\,\mx)} q_k(\mx,\Phi_k(\ma,\mc)) \sum_{\Phi_k^c(\ma,\mc)} \pi_0(\ma\,|\,\mx) \notag \\
    &= \sum_{\Phi_k(\ma,\mc)} \frac{\pi(\Phi_k(\ma,\mc)\,|\,\mx)}{\pi_0(\Phi_k(\ma,\mc)\,|\,\mx)} q_k(\mx,\Phi_k(\ma,\mc)) 
    \underbrace{\sum_{\Phi_k^c(\ma,\mc)} \pi_0(\Phi_k(\ma,\mc) \cup \Phi_k^c(\ma,\mc)\,|\,\mx)}_{= \pi_0(\Phi_k(\ma,\mc)\,|\,\mx)} \notag \\
    &= \sum_{\Phi_k(\ma,\mc)} \pi(\Phi_k(\ma,\mc)\,|\,\mx) q_k(\mx,\Phi_k(\ma,\mc)) \notag \\
    &= \sum_{\Phi_k(\ma,\mc)} q_k(\mx,\Phi_k(\ma,\mc)) \sum_{\Phi_k^c(\ma,\mc)} \pi(\Phi_k(\ma,\mc) \cup \Phi_k^c(\ma,\mc)\,|\,\mx) \notag \\
    &= \sum_{\ma} \pi(\ma\,|\,\mx) q_k(\mx,\ma,\mc) \notag \\
    &= \mE_{\pi(\ma|\mx)p(\mr|\mx,\ma,\mc)}[r_k] \label{eq:unbiased-final}
\end{align}
where $q_k(\mx,\Phi_k(\ma,\mc)) := \mE_{p(\mr|\mx,\Phi_k(\ma,\mc))}[r_k]$, $q_k(\mx,\ma,\mc) := \mE_{p(\mr|\mx,\ma,\mc)}[r_k]$, and $q_k(\mx,\ma,\mc) = q_k(\mx,\Phi_k(\ma,\mc))$.

Then, we have
\begin{align*}
    \mE_{\calD} \left[\hat{V}_k^{\mathrm{AIPS}}(\pi; \calD) \right]
    & = \mE_{\calD} \left[ \frac{1}{n} \sum_{i=1}^n \frac{\pi(\Phi_k(\ma_i,\mc_i)\,|\,\mx_i)}{\pi_0(\Phi_k(\ma_i,\mc_i)\,|\,\mx_i)} r_{i,k} \right] \\
    & =  \mE_{p(\mx)p(\mc|\mx)\pi_0(\ma|\mx)p(\mr|\mx,\ma,\mc)} \left[ \frac{\pi(\Phi_k(\ma,\mc)\,|\,\mx)}{\pi_0(\Phi_k(\ma,\mc)\,|\,\mx)} r_{k} \right]  \\
    & = \mE_{p(\mx)p(\mc|\mx)} \left[ \mE_{\pi_0(\ma|\mx)p(\mr|\mx,\ma,\mc)} \left[ \frac{\pi(\Phi_k(\ma,\mc)\,|\,\mx)}{\pi_0(\Phi_k(\ma,\mc)\,|\,\mx)} r_{k} \right] \right]  \\ 
    & = \mE_{p(\mx)p(\mc|\mx)} \left[ \mE_{\pi(\ma|\mx)p(\mr|\mx,\ma,\mc)}[r_k] \right] \quad \because \text{Eq.~\eqref{eq:unbiased-final}} \\ 
    & = V_k(\pi)
\end{align*}
\end{proof}

\subsection{Proof of Theorems~\ref{thrm:variance} and~\ref{thrm:minimum}}
\begin{proof} To prove Theorems~\ref{thrm:variance} and~\ref{thrm:minimum}, we quantify the difference between the variance of AIPS ($\hat{V}_k^{\mathrm{AIPS}}(\pi; \calD)$) and that of an arbitrary unbiased estimator ($\hat{V}_k(\pi; \calD, \tilde{\mc})$) defined in Theorem~\ref{thrm:minimum}. 
For brevity of exposition, the following uses $\Phi_k^d(\ma, \mc, \tmc) := \Phi_k(\ma, \tmc) \setminus \Phi_k(\ma, \mc)$. We will also use the fact that $\mc \subseteq \tilde{\mc}$ always holds true for any $\tmc$ when $\hat{V}_k(\pi; \calD, \tilde{\mc})$ is unbiased. This directly follows from Theorem~\ref{thrm:bias}, which is later proved in Appendix~\ref{app:proof_bias}
\begin{align}
    & n\left(\mV_{\calD}(\hat{V}_k(\pi; \calD, \tilde{\mc})) - \mV_{\calD}(\hat{V}_k^{\mathrm{AIPS}}(\pi; \calD)) \right) \notag \\
    &= n\left(\mV_{\calD} \left( \frac{\pi(\Phi_k(\ma,\tmc)\,|\,\mx)}{\pi_0(\Phi_k(\ma,\tmc)\,|\,\mx)} r_k \right) - \mV_{\calD} \left( \frac{\pi(\Phi_k(\ma,\mc)\,|\,\mx)}{\pi_0(\Phi_k(\ma,\mc)\,|\,\mx)} r_k \right) \right) \notag \\
    &= \mE_{p(\mx)p(\mc|\mx)\pi_0(\ma|\mx)p(\mr|\mx,\ma,\mc)}\left[ \left( \frac{\pi(\Phi_k(\ma,\tmc)\,|\,\mx)}{\pi_0(\Phi_k(\ma,\tmc)\,|\,\mx)} r_k \right)^2 \right] - \biggl(
    \underbrace{\mE_{p(\mx)p(\mc|\mx)\pi_0(\ma|\mx)p(\mr|\mx,\ma,\mc)}\left[ \frac{\pi(\Phi_k(\ma,\tmc)\,|\,\mx)}{\pi_0(\Phi_k(\ma,\tmc)\,|\,\mx)} r_k \right]}_{=V_k(\pi)} \biggr)^2 \notag \\
    & - \Biggl(\mE_{p(\mx)p(\mc|\mx)\pi_0(\ma|\mx)p(\mr|\mx,\ma,\mc)}\left[ \left( \frac{\pi(\Phi_k(\ma,\mc)\,|\,\mx)}{\pi_0(\Phi_k(\ma,\mc)\,|\,\mx)} r_k \right)^2 \right] - \biggl(\underbrace{\mE_{p(\mx)p(\mc|\mx)\pi_0(\ma|\mx)p(\mr|\mx,\ma,\mc)}\left[ \frac{\pi(\Phi_k(\ma,\mc)\,|\,\mx)}{\pi_0(\Phi_k(\ma,\mc)\,|\,\mx)} r_k \right]}_{=V_k(\pi)} \biggr)^2 \Biggr) \notag \\
    & = \mE_{p(\mx)p(\mc|\mx)\pi_0(\ma|\mx)p(\mr|\mx,\ma,\mc)}\left[  \left( \left( \frac{\pi(\Phi_k(\ma,\tmc)\,|\,\mx)}{\pi_0(\Phi_k(\ma,\tmc)\,|\,\mx)} \right)^2 - \left( \frac{\pi(\Phi_k(\ma,\mc)\,|\,\mx)}{\pi_0(\Phi_k(\ma,\mc)\,|\,\mx)} \right)^2 \right) r_k^2 \right] \notag \\
    & = \mE_{p(\mx)p(\mc|\mx)\pi_0(\ma|\mx)p(\mr|\mx,\ma,\mc)}\left[ \left( \frac{\pi(\Phi_k(\ma,\mc)\,|\,\mx)}{\pi_0(\Phi_k(\ma,\mc)\,|\,\mx)} \right)^2 \left( \left( \frac{\pi(\Phi_k(\ma,\tmc)\,|\,\mx)}{\pi_0(\Phi_k(\ma,\tmc)\,|\,\mx)} \frac{\pi_0(\Phi_k(\ma,\mc)\,|\,\mx)}{\pi(\Phi_k(\ma,\mc)\,|\,\mx)} \right)^2 - 1 \right) r_k^2\right] \notag \\
    &= \mE_{p(\mx)p(\mc|\mx)\pi_0(\ma|\mx)p(\mr|\mx,\ma,\mc)} \left[ \left( \frac{\pi(\Phi_k(\ma,\mc)\,|\,\mx)}{\pi_0(\Phi_k(\ma,\mc)\,|\,\mx)} \right)^2 
    \left( \left( \frac{\pi(\Phi_k^d(\ma, \mc, \tmc))\,|\,\mx,\Phi_k(\ma,\mc))}{\pi_0(\Phi_k^d(\ma, \mc, \tmc)\,|\,\mx,\Phi_k(\ma,\mc))} \right)^2 - 1 \right) r_k^2 \right] \notag \\
    &= \mE_{p(\mx)p(\mc|\mx)\pi_0(\ma|\mx)} \Biggl[ \left( \frac{\pi(\Phi_k(\ma,\mc)\,|\,\mx)}{\pi_0(\Phi_k(\ma,\mc)\,|\,\mx)} \right)^2  \Biggl( \Biggl( \frac{\pi(\Phi_k^d(\ma, \mc, \tmc)\,|\,\mx,\Phi_k(\ma,\mc))}{\pi_0(\Phi_k^d(\ma, \mc, \tmc)\,|\,\mx,\Phi_k(\ma,\mc))} \Biggr)^2 - 1 \Biggr) \,\, \mE_{p(\mr|\mx,\Phi_k(\ma,\mc))} \left[ r_k^2 \right] \Biggr] \notag \\
    &= \mE_{p(\mx)p(\mc|\mx)\pi_0(\Phi_k(\ma,\mc)|\mx)} \Biggl[ \left( \frac{\pi(\Phi_k(\ma,\mc)\,|\,\mx)}{\pi_0(\Phi_k(\ma,\mc)\,|\,\mx)} \right)^2  \mE_{\pi_0(\Phi_k^d(\ma, \mc, \tmc)\,|\,\mx,\Phi_k(\ma,\mc))} \Biggl[ \left( \frac{\pi(\Phi_k^d(\ma, \mc, \tmc)\,|\,\mx,\Phi_k(\ma,\mc))}{\pi_0(\Phi_k^d(\ma, \mc, \tmc)\,|\,\mx,\Phi_k(\ma,\mc))} \right)^2 - 1 \Biggr] \,\, \mE_{p(\mr|\mx,\Phi_k(\ma,\mc))} \left[ r_k^2 \right] \Biggr]  \notag \\
    &= \mE_{p(\mx)p(\mc|\mx)\pi_0(\Phi_k(\ma,\mc)|\mx)} \Biggl[ \left( \frac{\pi(\Phi_k(\ma,\mc)\,|\,\mx)}{\pi_0(\Phi_k(\ma,\mc)\,|\,\mx)} \right)^2 \mV_{\pi_0(\Phi_k^d(\ma, \mc, \tmc)\,|\,\mx,\Phi_k(\ma,\mc))}\left[ \frac{\pi(\Phi_k^d(\ma, \mc, \tmc)\,|\,\mx,\Phi_k(\ma,\mc))}{\pi_0(\Phi_k^d(\ma, \mc, \tmc)\,|\,\mx,\Phi_k(\ma,\mc))} \right] \, \mE_{p(\mr|\mx,\Phi_k(\ma,\mc))} \left[ r_k^2 \right] \Biggr], \label{eq:variance-reduction-final}
\end{align}
where we use $\frac{\pi(\Phi_k(\ma,\tmc)\,|\,\mx)}{\pi(\Phi_k(\ma,\mc)\,|\,\mx)} = \pi(\Phi_k^d(\ma, \mc, \tmc)\,|\,\mx,\Phi_k(\ma,\mc))$ and $\pi_0(\ma\,|\,\mx) = \pi_0(\Phi_k(\ma,\mc)\,|\,\mx) \pi_0(\Phi_k^d(\ma, \mc, \tmc)\,|\,\mx,\Phi_k(\ma,\mc))$. 
Moreover, in Eq.~\eqref{eq:variance-reduction-final}, we use the following trick:
\begin{align*}
    & \mE_{\pi_0(\Phi_k^d(\ma, \mc, \tmc)\,|\,\mx,\Phi_k(\ma,\mc))} \Biggl[ \left( \frac{\pi(\Phi_k^d(\ma, \mc, \tmc)\,|\,\mx,\Phi_k(\ma,\mc))}{\pi_0(\Phi_k^d(\ma, \mc, \tmc)\,|\,\mx,\Phi_k(\ma,\mc))} \right)^2 - 1 \Biggr] \\
    & = \mE_{\pi_0(\Phi_k^d(\ma, \mc, \tmc)\,|\,\mx,\Phi_k(\ma,\mc))} \Biggl[ \left( \frac{\pi(\Phi_k^d(\ma, \mc, \tmc)\,|\,\mx,\Phi_k(\ma,\mc))}{\pi_0(\Phi_k^d(\ma, \mc, \tmc)\,|\,\mx,\Phi_k(\ma,\mc))} \right)^2 \Biggr]  - \Biggl( \underbrace{\mE_{\pi_0(\Phi_k^d(\ma, \mc, \tmc)\,|\,\mx,\Phi_k(\ma,\mc))} \left[ \frac{\pi(\Phi_k^d(\ma, \mc, \tmc)\,|\,\mx,\Phi_k(\ma,\mc))}{\pi_0(\Phi_k^d(\ma, \mc, \tmc)\,|\,\mx,\Phi_k(\ma,\mc))} \right]}_{=1} \Biggr)^2 \\
    & = \mV_{\pi_0(\Phi_k^d(\ma, \mc, \tmc)\,|\,\mx,\Phi_k(\ma,\mc))}\left[ \frac{\pi(\Phi_k^d(\ma, \mc, \tmc)\,|\,\mx,\Phi_k(\ma,\mc))}{\pi_0(\Phi_k^d(\ma, \mc, \tmc)\,|\,\mx,\Phi_k(\ma,\mc))} \right]
\end{align*}
We can see that Eq.~\eqref{eq:variance-reduction-final} is always non-negative, which means that the variance of AIPS is never larger than that of any unbiased IPS estimator defined by $\hat{V}_k(\pi; \calD, \tilde{\mc})$ with $\mc \subseteq \tilde{\mc}$. Hence, Theorem~\ref{thrm:minimum} is proved. Furthermore, we can derive Theorem~\ref{thrm:variance} by replacing $\Phi_k(\ma,\tmc)$ with $\ma$ (in this case, $\Phi_k^d(\ma, \mc, \tmc) = \Phi_k^c(\ma,\mc)$).
\end{proof}

\subsection{Proof of Theorem~\ref{thrm:bias}} \label{app:proof_bias}

\begin{proof}
First, we calculate the bias of AIPS with an estimated user behavior $\hat{\mc}$ below.

For any given $\mx \sim p(\mx)$ and $\mc \sim p(\mc\,|\,\mx)$, we have
\begin{align}
    & \mE_{\pi_0(\ma|\mx)p(\mr|\mx,\ma,\mc)} \left[ \frac{\pi(\Phi_k(\ma,\hat{\mc})\,|\,\mx)}{\pi_0(\Phi_k(\ma,\hat{\mc})\,|\,\mx)} r_k \right]  \notag \\
    &= \sum_{\ma} \pi_0(\ma\,|\,\mx) \frac{\pi(\Phi_k(\ma,\hat{\mc})\,|\,\mx)}{\pi_0(\Phi_k(\ma,\hat{\mc})\,|\,\mx)} q_k(\mx,\ma,\mc) \notag \\
    &= \sum_{\Phi_k(\ma,\mc) \cup \Phi_k(\ma,\hat{\mc})} \sum_{\Phi_k^c(\ma,\mc) \cap \Phi_k^c(\ma,\hat{\mc})} \pi_0(\ma\,|\,\mx) \frac{\pi(\Phi_k(\ma,\hat{\mc})\,|\,\mx)}{\pi_0(\Phi_k(\ma,\hat{\mc})\,|\,\mx)}  q_k(\mx,\ma,\mc) \notag \\
    &= \sum_{\Phi_k(\ma,\mc) \cup \Phi_k(\ma,\hat{\mc})} \frac{\pi(\Phi_k(\ma,\hat{\mc})\,|\,\mx)}{\pi_0(\Phi_k(\ma,\hat{\mc})\,|\,\mx)} 
    q_k(\mx,\Phi_k(\ma,\mc)) \sum_{\Phi_k^c(\ma,\mc) \cap \Phi_k^c(\ma,\hat{\mc})} \pi_0(\ma\,|\,\mx) \notag \\
    &= \sum_{\Phi_k(\ma,\mc) \cup \Phi_k(\ma,\hat{\mc})} \frac{\pi(\Phi_k(\ma,\hat{\mc})\,|\,\mx)}{\pi_0(\Phi_k(\ma,\hat{\mc})\,|\,\mx)} q_k(\mx,\Phi_k(\ma,\mc)) \underbrace{\sum_{\Phi_k^c(\ma,\mc) \cap \Phi_k^c(\ma,\hat{\mc})} \pi_0(\big(\Phi_k(\ma,\mc) \cup \Phi_k(\ma,\hat{\mc})\big) \cup \big(\Phi_k^c(\ma,\mc) \cap \Phi_k^c(\ma,\hat{\mc})\big)\,|\,\mx)}_{=\pi_0(\Phi_k(\ma,\mc) \cup \Phi_k(\ma,\hat{\mc})\,|\,\mx)} \notag \\
    &= \sum_{\Phi_k(\ma,\mc) \cup \Phi_k(\ma,\hat{\mc})} \frac{\pi(\Phi_k(\ma,\hat{\mc})\,|\,\mx)}{\pi_0(\Phi_k(\ma,\hat{\mc})\,|\,\mx)} q_k(\mx,\Phi_k(\ma,\mc)) \frac{\pi_0(\Phi_k(\ma,\mc) \cup \Phi_k(\ma,\hat{\mc})\,|\,\mx)}{\pi(\Phi_k(\ma,\mc) \cup \Phi_k(\ma,\hat{\mc})\,|\,\mx)} \pi(\Phi_k(\ma,\mc) \cup \Phi_k(\ma,\hat{\mc})\,|\,\mx) \notag \\
    &= \sum_{\Phi_k(\ma,\mc) \cup \Phi_k(\ma,\hat{\mc})} \frac{\pi(\Phi_k(\ma,\hat{\mc})\,|\,\mx)}{\pi_0(\Phi_k(\ma,\hat{\mc})\,|\,\mx)} q_k(\mx,\Phi_k(\ma,\mc)) \frac{\pi_0(\Phi_k(\ma,\hat{\mc})\,|\,\mx)\pi_0(\Phi_k(\ma,\mc) \setminus \Phi_k(\ma,\hat{\mc}))\,|\,\mx,\Phi_k(\ma,\hat{\mc})) }{\pi(\Phi_k(\ma,\hat{\mc})\,|\,\mx)\pi(\Phi_k(\ma,\mc) \setminus \Phi_k(\ma,\hat{\mc}))\,|\,\mx,\Phi_k(\ma,\hat{\mc}))} \pi(\Phi_k(\ma,\mc) \cup \Phi_k(\ma,\hat{\mc})\,|\,\mx) \notag \\
    &= \sum_{\Phi_k(\ma,\mc) \cup \Phi_k(\ma,\hat{\mc})} \frac{\pi_0(\Phi_k(\ma,\mc) \setminus \Phi_k(\ma,\hat{\mc}))\,|\,\mx,\Phi_k(\ma,\hat{\mc}))}{\pi(\Phi_k(\ma,\mc) \setminus \Phi_k(\ma,\hat{\mc}))\,|\,\mx,\Phi_k(\ma,\hat{\mc}))}  q_k(\mx,\Phi_k(\ma,\mc)) \pi(\Phi_k(\ma,\mc) \cup \Phi_k(\ma,\hat{\mc})\,|\,\mx) \notag \\
    &= \sum_{\Phi_k(\ma,\mc) \cup \Phi_k(\ma,\hat{\mc})} \frac{\pi_0(\Phi_k(\ma,\mc) \setminus \Phi_k(\ma,\hat{\mc}))\,|\,\mx,\Phi_k(\ma,\hat{\mc}))}{\pi(\Phi_k(\ma,\mc) \setminus \Phi_k(\ma,\hat{\mc}))\,|\,\mx,\Phi_k(\ma,\hat{\mc}))}  q_k(\mx,\Phi_k(\ma,\mc)) \sum_{\Phi_k^c(\ma,\mc) \cap \Phi_k^c(\ma,\hat{\mc})} \pi(\big(\Phi_k(\ma,\mc) \cup \Phi_k(\ma,\hat{\mc})\big) \cup \big(\Phi_k^c(\ma,\mc) \cap \Phi_k^c(\ma,\hat{\mc})\big)\,|\,\mx) \notag \\
    &= \sum_{\ma} \pi(\ma\,|\,\mx) \frac{\pi_0(\Phi_k(\ma,\mc) \setminus \Phi_k(\ma,\hat{\mc}))\,|\,\mx,\Phi_k(\ma,\hat{\mc}))}{\pi(\Phi_k(\ma,\mc) \setminus \Phi_k(\ma,\hat{\mc}))\,|\,\mx,\Phi_k(\ma,\hat{\mc}))} q_k(\mx,\ma,\mc) \notag \\
    &= \mE_{\pi(\ma|\mx)} \left[ \frac{\pi_0(\Phi_k(\ma,\mc) \setminus \Phi_k(\ma,\hat{\mc}))\,|\,\mx,\Phi_k(\ma,\hat{\mc}))}{\pi(\Phi_k(\ma,\mc) \setminus \Phi_k(\ma,\hat{\mc}))\,|\,\mx,\Phi_k(\ma,\hat{\mc}))}  q_k(\mx,\ma,\mc) \right] \label{eq:bias-final}
\end{align}

Therefore, we have
\begin{align*}
    \mathrm{Bias}(\hat{V}_k^{\mathrm{AIPS}}; \hat{\mc})
    & = \mE_{\calD} \left[ \frac{1}{n} \sum_{i=1}^n \frac{\pi(\Phi_k(\ma_i,\hat{\mc}_i)\,|\,\mx_i)}{\pi_0(\Phi_k(\ma_i,\hat{\mc}_i)\,|\,\mx_i)} r_{i,k} \right] - V_k(\pi) \\
    & =  \mE_{p(\mx)p(\mc|\mx)\pi_0(\ma|\mx)p(\mr|\mx,\ma,\mc)} \left[ \frac{\pi(\Phi_k(\ma,\hat{\mc})\,|\,\mx)}{\pi_0(\Phi_k(\ma,\hat{\mc})\,|\,\mx)} r_{k} \right] - \mE_{p(\mx)p(\mc|\mx)\pi(\ma|\mx)p(\mr|\mx,\ma,\mc)} [r_k]  \\
    & = \mE_{p(\mx)p(\mc|\mx)} \left[ \mE_{\pi_0(\ma|\mx)p(\mr|\mx,\ma,\mc)} \left[ \frac{\pi(\Phi_k(\ma,\hat{\mc})\,|\,\mx)}{\pi_0(\Phi_k(\ma,\hat{\mc})\,|\,\mx)} r_{k} \right] \right]  -  \mE_{p(\mx)p(\mc|\mx)\pi(\ma|\mx)} [q_k(\mx,\ma,\mc)] \\ 
    & = \mE_{p(\mx)p(\mc|\mx)\pi(\ma|\mx)} \left[ \frac{\pi_0(\Phi_k(\ma,\mc) \setminus \Phi_k(\ma,\hat{\mc}))\,|\,\mx,\Phi_k(\ma,\hat{\mc}))}{\pi(\Phi_k(\ma,\mc) \setminus \Phi_k(\ma,\hat{\mc}))\,|\,\mx,\Phi_k(\ma,\hat{\mc}))}  q_k(\mx,\ma,\mc) \right] - \mE_{p(\mx)p(\mc|\mx)\pi(\ma|\mx)} [q_k(\mx,\ma,\mc)] \quad \because \text{Eq.~\eqref{eq:bias-final}} \\ 
    & = \mE_{p(\mx)p(\mc|\mx)\pi(\ma|\mx)} \left[ \left(\frac{\pi_0(\Phi_k(\ma,\mc) \setminus \Phi_k(\ma,\hat{\mc}))\,|\,\mx,\Phi_k(\ma,\hat{\mc}))}{\pi(\Phi_k(\ma,\mc) \setminus \Phi_k(\ma,\hat{\mc}))\,|\,\mx,\Phi_k(\ma,\hat{\mc}))} - 1 \right) q_k(\mx,\ma,\mc) \right]  \\ 
    & = \mE_{p(\mx)p(\mc|\mx)\pi(\ma|\mx)} \left[ \left( \Delta w_k(\ma, \mc, \hat{\mc}) - 1 \right)  q_k(\mx,\ma,\mc) \right]
\end{align*}
where 
\begin{align*}
    \Delta w_k(\ma, \mc, \hat{\mc}) := \frac{\pi_0(\Phi_k(\ma,\mc) \setminus \Phi_k(\ma,\hat{\mc}) \,|\, \mx,\Phi_k(\ma,\hat{\mc}))}{\pi(\Phi_k(\ma,\mc) \setminus \Phi_k(\ma,\hat{\mc}) \,|\, \mx,\Phi_k(\ma,\hat{\mc}))}.
\end{align*}
\end{proof}

\begin{figure*}[tb]
\begin{minipage}[b]{0.8\linewidth}
\centering
\begin{algorithm}[H]
\caption{The procedure to optimize user behavior assignments in AIPS (detailed in Section~\ref{sec:ubso})} \label{algo:ubsi} 
  \begin{algorithmic}[1]
    \REQUIRE logged data $\calD$, a loss function to minimize MSE $\mathcal{L}(\cdot)$, a set of candidate user behavior models $\mathcal{C} = \{ \hat{\mc}^0, \cdots, \hat{\mc}^{m} \}$, the base user behavior model $\hat{\mc}_{base}$, a set of random states $\mathcal{S}$
    \ENSURE dictionary containing $\hat{\mc}$ for each partition $\mathbb{C}$
    \STATE Initialize node sets to partition $\mathbb{L} \leftarrow \{ \calD \} $ 
    and dictionary containing $\hat{\mc}$ for each partition $\mathbb{C} \leftarrow \emptyset$
    \STATE Initialize the number of user partition $g \leftarrow 0$
    \WHILE{$\mathbb{L} \neq \emptyset$}
        \STATE Remove node $l$ from $\mathbb{L}$ as $\mathbb{L} \leftarrow \mathbb{L} \setminus \{ l \}$ and set $l$ as the parent node 
        \STATE Initialize the minimum loss $\widehat{MSE}^{(-)} \leftarrow \widehat{MSE} (\hat{\mc}_{(l)}; l)$ where $ \hat{\mc}_{(l)} := \argmin_{\hat{\mc}} \widehat{MSE}(\hat{\mc}; l)$
        \STATE Initialize the best subset  $(\hat{c}_{(l_{(l)})}, \hat{c}_{(l_{(r)})}, \calD_{(l_{(l)})}, \calD_{(l_{(r)})})
        \leftarrow \emptyset$
        \FOR{$s \in \mathcal{S}$}
            \STATE Randomly generate partitions in the feature space ($\calX$) and create two subsets of the data $(\calD_{(l^{\ast})}, \calD_{(r^{\ast})})$ \\
            \quad (e.g., data that satisfy $||\mx||_2 \leq 1$ are deemed as the subset indicating the left node ($\calD_{(l^{\ast})}$), while others \\
            \quad as the subset of the right node ($\calD_{(r^{\ast})}$))
            \STATE Identify the best behavior model for each subset as\\
            $\quad (\hat{\mc}_{(l^{\ast})}, \hat{\mc}_{(r^{\ast})}) 
            := \argmin_{(\hat{\mc}_{(l)}, \hat{\mc}_{(r)})} \, \widehat{MSE}(\hat{\mc}_{(l)}, \hat{\mc}_{(r)}; l^{\ast}, r^{\ast})$ \\
            \IF{$\widehat{MSE}^{(-)} > \widehat{MSE}(\hat{\mc}_{(l^{\ast})}, \hat{\mc}_{(r^{\ast})}; l^{\ast}, r^{\ast})$}
                \STATE Update the best partition as \\
                $\quad \widehat{MSE}^{(-)} \leftarrow \widehat{MSE}(\hat{\mc}_{(l^{\ast})}, \hat{\mc}_{(r^{\ast})}; l^{\ast}, r^{\ast})$\\ $\quad (\hat{c}_{(l_{(l)})}, \hat{c}_{(l_{(r)})}, \calD_{(l_{(l)})}, \calD_{(l_{(r)})})
                \leftarrow (\hat{c}_{(l^{\ast})}, \hat{c}_{(r^{\ast})}, \calD_{(l^{\ast})}, \calD_{(r^{\ast})})$
            \ENDIF
        \ENDFOR
        \IF{$(\calD^{(l_{(l)})}, \calD^{(l_{(r)})}) = \emptyset$} 
            \STATE // end of the optimization procedure
            \STATE Add the parent partition and the corresponding user behavior model to $\mathbb{C}$ as \\
            $\quad \mathbb{C}[g] \leftarrow (\calD_{(l)}, \hat{\mc}_{(l)}, l)$, $\quad g \leftarrow g + 1$
        \ELSE
            \STATE // continue the optimization procedure
            \STATE Add children nodes to the tree as 
            $\; \mathbb{L} \leftarrow \mathbb{L} \bigcup \{ \calD_{(l_{(l)})}, \calD_{(l_{(r)})} \} $
        \ENDIF
    \ENDWHILE
  \end{algorithmic}
\end{algorithm}
\end{minipage}
\end{figure*}